\documentclass[letterpaper]{article} 
\usepackage{aaai2026}  
\usepackage{times}  
\usepackage{helvet}  
\usepackage{courier}  
\usepackage[hyphens]{url}  
\usepackage{graphicx} 
\usepackage{natbib}  
\usepackage{caption} 
\usepackage{graphicx}
\usepackage{amsmath}
\usepackage{amssymb}
\usepackage{etoolbox}
\usepackage{multirow}
\usepackage{algorithm}
\usepackage{enumitem}
\usepackage{booktabs}
\usepackage{xcolor}
\usepackage{xspace} 

\usepackage{newfloat}
\usepackage{listings}
\DeclareCaptionStyle{ruled}{labelfont=normalfont,labelsep=colon,strut=off} 
\floatstyle{ruled}
\newfloat{listing}{tb}{lst}{}
\floatname{listing}{Listing}
\newcommand{\method}{IdentityStory\xspace}  
\newcommand{\techa}{Iterative Identity Discovery\xspace}
\newcommand{\techb}{Re-denoising Identity Injection\xspace}

\newcommand{\myparagraph}[1]{\vspace{0.3em}\noindent\textbf{#1}}
\newcommand{\eg}{\textit{e.g.}\@\xspace}

\title{\method: Taming Your Identity-Preserving Generator for \\ Human-Centric Story Generation}

\author {
    \large
    Donghao Zhou\textsuperscript{\rm 1*},
    Jingyu Lin\textsuperscript{\rm 2*},
    Guibao Shen\textsuperscript{\rm 3},
    Quande Liu\textsuperscript{\rm 4},
    Jialin Gao\textsuperscript{\rm 1},
    Lihao Liu\textsuperscript{\rm 5},
    Lan Du\textsuperscript{\rm 2}, \\
    Cunjian Chen\textsuperscript{\rm 2$\dagger$},
    Chi-Wing Fu\textsuperscript{\rm 1}, 
    Xiaowei Hu\textsuperscript{\rm 6$\dagger$},
    Pheng-Ann Heng\textsuperscript{\rm 1} \vspace{1mm}
}
\affiliations {
    \textsuperscript{\rm 1}The Chinese University of Hong Kong, \quad
    \textsuperscript{\rm 2}Monash University, \\
    \textsuperscript{\rm 3}The Hong Kong University of Science and Technology (Guangzhou), \quad
    \textsuperscript{\rm 4}Kling Team, Kuaishou Technology, \\
    \textsuperscript{\rm 5}Amazon, \quad
    \textsuperscript{\rm 6}South China University of Technology\\
    \vspace{1mm} {\small \texttt{Project Page:}} {\small \urlstyle{tt}\textcolor{purple}{\url{https://correr-zhou.github.io/IdentityStory}}}
}

\begin{document}

\maketitle

\let\thefootnote\relax\footnotetext{\\
\textsuperscript{\rm *}Equal contribution. \\ \textsuperscript{\rm $\dagger$}Corresponding authors.
}

\begin{abstract}

Recent visual generative models enable story generation with consistent characters from text, but human-centric story generation faces additional challenges, such as maintaining detailed and diverse human face consistency and coordinating multiple characters across different images.
This paper presents \textbf{IdentityStory}, a framework for human-centric story generation that ensures consistent character identity across multiple sequential images. By taming identity-preserving generators, the framework features two key components: \textit{Iterative Identity Discovery}, which extracts cohesive character identities, and \textit{Re-denoising Identity Injection}, which re-denoises images to inject identities while preserving desired context. Experiments on the ConsiStory-Human benchmark demonstrate that IdentityStory outperforms existing methods, particularly in face consistency, and supports multi-character combinations. The framework also shows strong potential for applications such as infinite-length story generation and dynamic character composition.

\end{abstract}


\section{Introduction}
\label{sec:intro}

Recent visual generative models \cite{rombach2022high, podell2023sdxl, chen2023pixart, FLUX} enable users to create high-quality images from text, but they still struggle to maintain character consistency across multiple generated images due to their stochastic nature.
This limitation promotes the task of story generation, which aims to yield a series of images with consistent characters solely using text.
While this task has already made an impact in education \cite{carter1993place} and entertainment \cite{klimmt2012forecasting}, it exhibits greater potential in human-centric scenarios such as film storyboarding \cite{hart2013art, halligan2013movie}, advertisement design \cite{escalas2003advertising, megehee2010creating}, and artistic production \cite{cetinic2022understanding}.
Therefore, we aim to further advance \textbf{human-centric story generation} in this work, exploring story generation specifically with humans as characters.

\begin{figure}[t]

    \centering
    
    
    \includegraphics[width=1\linewidth]{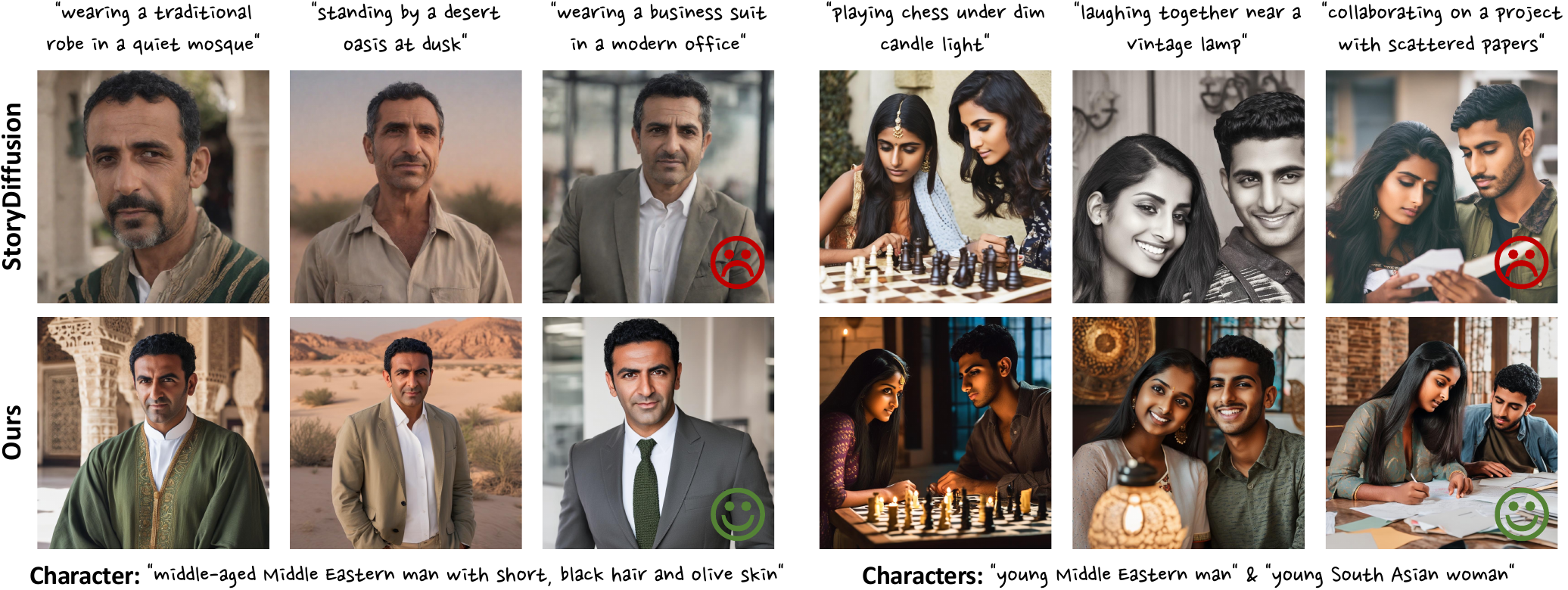}
    
    \vspace{-2mm}
    \captionsetup{hypcap=false}
    \captionof{figure}
    { 
        \textbf{Human-centric story generation.}
       Our \method can \textit{solely rely on text} to generate a series of images that consistently depict human characters and faithfully align with text prompts, outperforming the state-of-the-art method \cite{StoryDiffusion}. \textit{Zoom in for better view.}
    }
    \vspace{-4mm}
    
    \label{fig:teaser}
    
\end{figure}

\begin{figure}[t]
    \centering

    \includegraphics[width=1\linewidth]{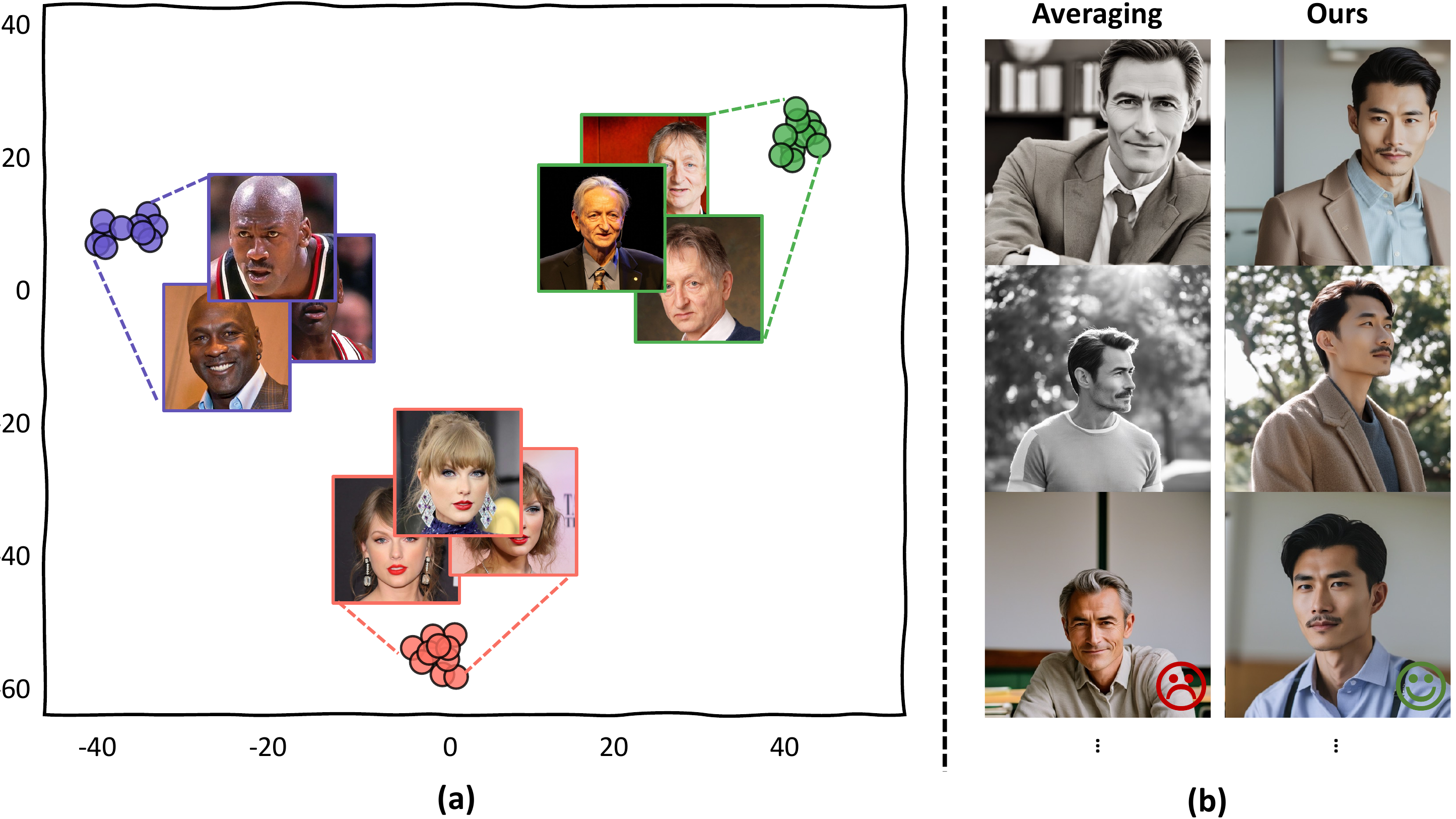}

    \vspace{-3mm}
    \caption
    {
        \textbf{Identity embeddings of identity-preserving generators.} 
        Here we use PhotoMaker~\cite{li2024photomaker} as an example:
        (a) We extract embeddings of three identities using Photomaker's image encoder and visualize them in 2D with t-SNE \cite{van2008visualizing}, observing that \textit{embeddings of different identities are highly distinguishable}.
        (b) 
        We collect 20 character descriptions and extract identity embeddings using the naive averaging approach and our method.
        Then, each embedding generates 15 images for computing pairwise face similarity with ArcFace~\cite{deng2019arcface}, showing our method performs better.
    }
    \vspace{-2mm}
    
    \label{fig:motivation_emb}
    
\end{figure}

As shown in Figure~\ref{fig:teaser}, this task adopts a text prompt set where each prompt shares the same character descriptions as input, aiming to generate multiple images with consistent characters and the corresponding visual content. 
Human-centric story generation is more challenging compared to story generation on other subjects, since  
(i) human faces inherently contain richer details \cite{wang2021deep} and exhibit diverse variations \cite{zhou2021face, kammoun2022generative}, imposing greater difficulty on consistency maintenance, and 
(ii) the coordination of multiple characters across different images demands a more flexible generation framework. 
Existing state-of-the-art methods of story generation cannot achieve satisfactory results (Figure~\ref{fig:teaser}\&\ref{fig:qual_results}), as they rely on attention-sharing mechanisms (\eg, ConsiStory~\cite{Consistency}, Story Diffusion~\cite{StoryDiffusion}) or global semantic modulation (\eg, Story-Adapter~\cite{mao2024story}, 1Prompt1Story~\cite{liu2025one}), lacking the ability to precisely and flexibly maintain the cross-image identity consistency of human characters.

Recently, identity-preserving generators \cite{li2024photomaker, ip-adapter, wang2024instantid} have emerged within the task of subject-driven generation (requires reference images as input) \cite{gal2022image, ruiz2023dreambooth, rout2024rb, zhou2024magictailor} to ensure identity preservation in generated images.
Therefore, an intriguing question arises: \textit{Can we harness identity-preserving generators for human-centric story generation?}
In this work, we present \textbf{\method}, a framework designed to unleash the full potential of identity-preserving generators for this task, ensuring precise and flexible consistency maintenance for human characters across multiple images (Figure~\ref{fig:teaser}).

\begin{figure}[t]
    \centering

    \includegraphics[width=1\linewidth]{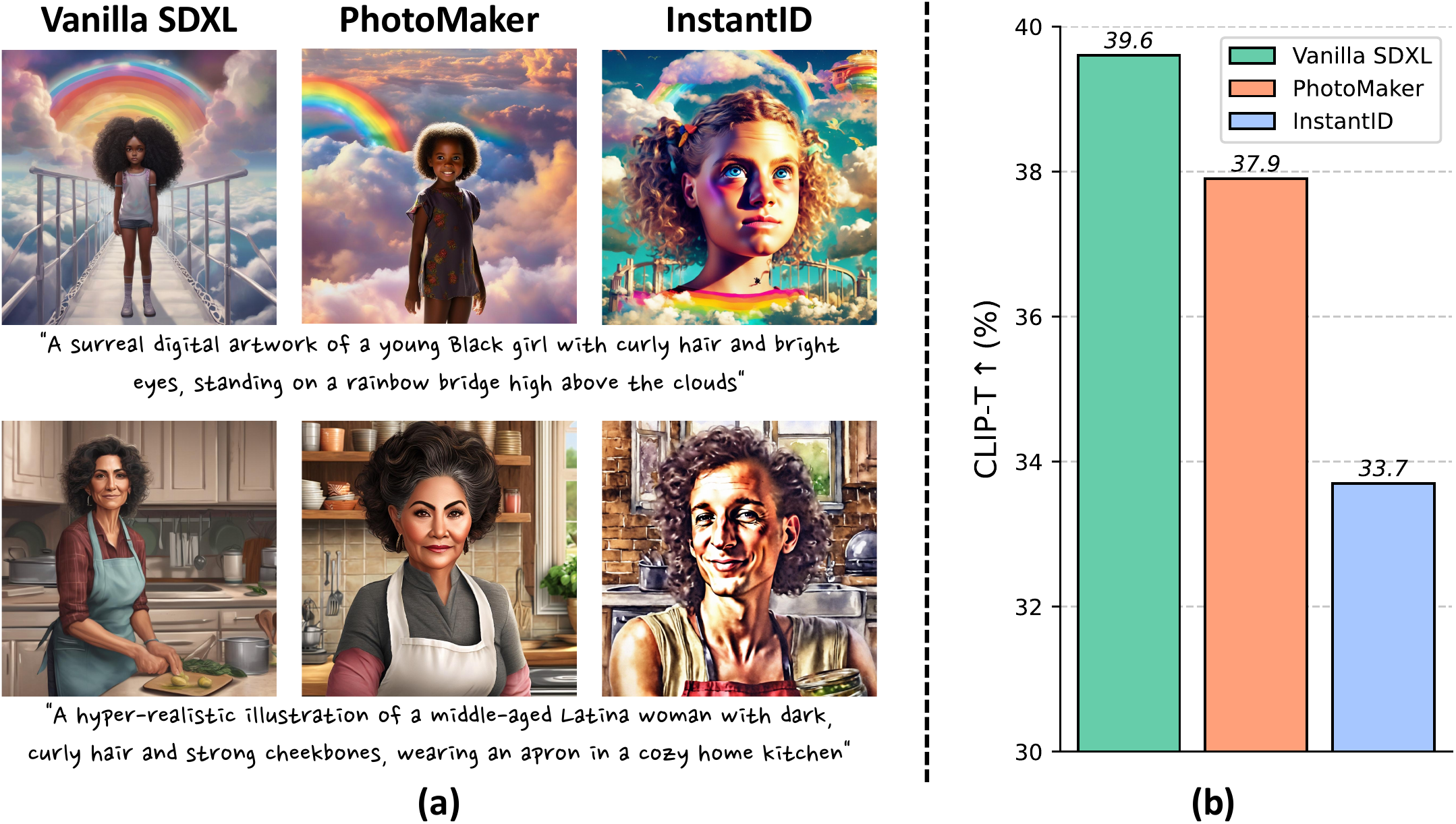}

    \vspace{-3mm}
    \caption
    {
        \textbf{Text alignment degradation of identity-preserving generators.}
        We collect 1,000 human-centric text prompts to compare the text alignment performance of vanilla SDXL~\cite{podell2023sdxl}, PhotoMaker~\cite{li2024photomaker}, and InstantID~\cite{wang2024instantid}. We present (a) qualitative and (b) quantitative results with the average CLIP score (CLIP-T) \cite{radford2021learning}.
        For the identity-preserving generators, we use SDXL as the base model and generate a human image as the reference.
    }
    \vspace{-2mm}
    
    \label{fig:motivation_align}
         
\end{figure}

\begin{figure*}[t]
    \centering
    
    \includegraphics[width=0.97\linewidth]{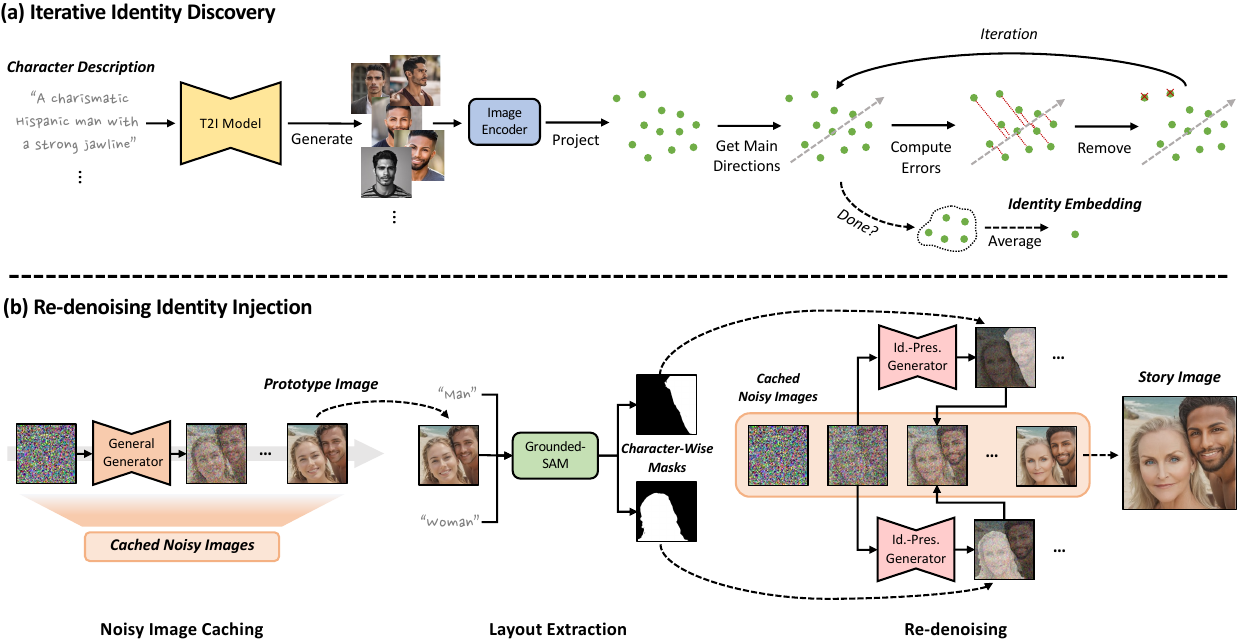}
    
    \vspace{-2mm}
    \caption
    {
        \textbf{Pipeline of \method.}
        This framework consists of two key techniques, including
        (a) \textit{\techa} (Section~\ref{sec:techa}), which utilizes Singular Value Decomposition (SVD) to iteratively filter out low-relevance embeddings to extract cohesive identities,
        and (b) \textit{\techb} (Section~\ref{sec:techb}), which uses identity-preserving generators to inject extracted identities based on the noisy cached images and character layouts produced from general generators.
    }
    \vspace{-2mm}
    
    \label{fig:pipeline}  
    
\end{figure*}

To achieve this, we first develop \textit{\techa} to extract identities (Figure~\ref{fig:pipeline}(a)). 
We find that identity-preserving generators process a well-constructed identity space (Figure~\ref{fig:motivation_emb}(a)), where identity representation can be obtained by aggregating character image embeddings. 
After generating diverse character images from descriptions and projecting them into the identity space, we use Singular Value Decomposition (SVD) to iteratively filter out low-relevance embeddings and extract cohesive identities.

Next, we design \textit{\techb} to inject identities (Figure~\ref{fig:pipeline}(b)). To address text alignment degradation of identity-preserving generators (Figure~\ref{fig:motivation_align}), we first use a general generator to create a more text-aligned prototype image. 
Meanwhile, we cache noisy images during generation to preserve environmental semantics and segment the prototype image to extract character layouts. 
Using a progressive masking strategy, we then re-denoise with identity-preserving generators to inject identities.

We perform experiments based on \textit{ConsiStory-Human}, a new benchmark improved from \citep{Consistency}, demonstrating that \method achieves overall superior performance and especially excels at precise face consistency maintenance.
Notably, our method achieves a pairwise face similarity score of 55.5\%, over double that of the second-best method (27.1\%).
Due to the decoupled nature of identity extraction and injection, our method also supports multi-character combinations across different images. 
Moreover, we show the practicality of \method to more applications such as community tool integration, infinite-length story generation, 
and dynamic character composition.

\section{Related Works}
\label{sec:related}

\noindent \textbf{Story Generation}
is first formulated as the task of story visualization by StoryGAN \cite{li2019storygan} and has evolved across diverse technical paradigms such as generative adversarial networks~\cite{li2019storygan,li2020improved}, large language-vision models \cite{shen2023large, yang2024seed}, and diffusion models~\cite{Consistency, StoryDiffusion,  mao2024story, liu2025one}.
Early works~\cite{li2019storygan,li2020improved,maharana2022storydall} tend to train models on close-domain datasets. 
While these methods manage to reproduce specified characters, the lack of generalization severely hinders their ability to handle unseen subjects or adapt to various contexts.
Building upon text-to-image (T2I) diffusion models, subsequent works~\cite{Consistency, StoryDiffusion, mao2024story, liu2025one} succeed in generating consistent characters of open domains.
However, these approaches involve either intricate module assembly \cite{gong2023talecrafter,liu2024intelligent}, complex strategy designs \cite{mao2024story, liu2025one}, or heavy memory usage \cite{Consistency, StoryDiffusion, mao2024story}, thus restricting their applicability in practice.
More importantly, they failed to precisely yield consistent human characters, while our \method excels in this aspect by unlocking the capabilities of identity-preserving generators.

\myparagraph{Subject-Driven Generation} aims to enable T2I diffusion models to generate specific visual concepts from reference images.
Initial approaches \cite{gal2022image, ruiz2023dreambooth} achieved this by optimizing text embeddings or fine-tuning model weights.
Subsequent methods \cite{kumari2023multi, jiang2024mc, kong2024omg, zhou2024magictailor} expanded on these approaches to handle multiple visual concepts.
Recently, learning-based approaches \cite{wei2023elite, he2024disenvisioner, tan2024ominicontrol, cai2024diffusion} have gained widespread adoption due to their strong zero-shot capabilities. 
Within these, a subset of methods \cite{li2024photomaker, wang2024instantid, xiao2024fastcomposer, wang2024stableidentity, guo2025pulid}, which we refer to as identity-preserving generators, specifically focus on generating human-centric images. 
With additional training on large human datasets, these methods effectively preserve the identity of individuals depicted in reference images \cite{wei2025personalized}.
In this work, our method unlocks their potential and goes beyond their original functionality, enabling them to support text-only input and achieve better text alignment.

\section{Methodology}
\label{sec:method}

In this work, we explore \textit{human-centric story generation}, aiming to generate visually coherent story images with human characters (Figure~\ref{fig:teaser}).
Specifically, the input is a story text $\mathcal{T} = \{T_i\}_{i=1}^n$ with $n$ text prompts, where each text prompt $T_i$ corresponds to a story image and shares the same character descriptions $\{c_i\}_{i=1}^s$ with $s$ characters. 
The goal is to generate a series of images $\mathcal{I} = \{I_i\}_{i=1}^n$ with consistent characters solely using $\mathcal{T} = \{T_i\}_{i=1}^n$.
We develop \textbf{\method}, aiming to unleash the potential of identity-preserving generators for human-centric story generation (Figure~\ref{fig:pipeline}).
In this framework,
\textit{\techa} (Sec.~\ref{sec:techa}) is first conducted to extract identity embeddings solely from character descriptions via iterative filtering,
and then \textit{\techb} (Sec.~\ref{sec:techb}) is employed to generate story images by injecting identity embeddings with a re-denoising paradigm.
In the following, we delve deeper into the details of these two techniques.

\subsection{\techa}
\label{sec:techa}

We observe that identity-preserving generators contain a well-structured identity space (Figure~\ref{fig:motivation_emb}(a)), making it possible to obtain an identity embedding by aggregating the embeddings of generated character images.
However, simply averaging the embeddings results in low identity uniqueness (Figure~\ref{fig:motivation_emb}(b)) due to the diversity of generated images, which reflect various identities.
Therefore, we develop \textit{\techa} (Figure~\ref{fig:pipeline}(a)), which leverages Singular Value Decomposition (SVD) to iteratively filter out low-relevance embeddings and ultimately aggregate the most cohesive ones, yielding a more unique identity representation (Figure~\ref{fig:motivation_emb}(b)).
Below we describe it in detail.

\myparagraph{Overall Scheme.}
We first generate $m$ character images for each character description $c_i$.
Using the image encoder of the identity-preserving generator $G_{\text{id}}$, these images are then projected into the identity space,  which results in a combined embedding matrix $E \in \mathbb{R}^{m \times d}$ where $d$ is the embedding dimension.
Inspired by \cite{gu2014weighted}, we obtain the implicit semantic information of $E$ by applying SVD as
\begin{equation}
    E = U \Sigma V^{T},
    \label{eq:svd}
\end{equation}
where $U \in \mathbb{R}^{m \times m}$,\ $\Sigma \in \mathbb{R}^{m \times d}$, and $V \in \mathbb{R}^{d \times d}$.
As a result, $V$ contains right singular vectors of $E$, where each column denotes a semantic direction of the identity space.
Next, we select the directions with the top $k$ singular values from $V$, formulating $V_k \in \mathbb{R}^{d \times k}$.
The directions in $V_k$ capture the most significant semantics of the embeddings, while the others involve low-relevance semantics or noises.
Then, we further compute the reconstruction matrix
$
    W = V_k V_k^T,
$
where $W \in \mathbb{R}^{d \times d}$ can be regarded as a transformation designed to preserve only the most essential semantics.
Using $W$, we can calculate the reconstruction errors $\varepsilon \in \mathbb{R}^{m}$ for $m$ identity embeddings as
\begin{equation}
    \varepsilon = \frac{1}{d} \sum_{i=1}^{d} \| E_{:,i} - (EW)_{:,i} \|_2,
\end{equation}
where $\{:,i\}$ represents the matrix indexing.
Low-relevance embeddings can deviate from the core semantics, thereby leading to larger errors due to insufficient reconstruction.
With a filtering ratio $r$ and an iteration number $p$, we remove the embeddings with the $(1-r)\cdot m$ largest errors from $E$, and return to Equation~\ref{eq:svd} for the next iteration (more details in Algorithm~\ref{algo:techa}).
Through this iterative process, about 21.6\% of embeddings are retained and averaged to form the final identity embedding, effectively capturing the core semantics of the character while minimizing the influence of non-identity factors such as poses and facial expressions.
Such a technique can extract a more reliable identity representation ensuring character consistency.

\begin{algorithm}[t]
\caption{Pseudocode of \techa.}
\label{algo:techa}
\definecolor{codeblue}{rgb}{0.25,0.5,0.5}
\lstset{
  backgroundcolor=\color{white},
  basicstyle=\fontsize{7.2pt}{7.2pt}\ttfamily\selectfont,
  columns=fullflexible,
  breaklines=true,
  captionpos=b,
  commentstyle=\fontsize{7.2pt}{7.2pt}\color{codeblue},
  keywordstyle=\fontsize{7.2pt}{7.2pt},
}

\begin{lstlisting}[language=python]
# The pseudocode is presented in the PyTorch style.
# Input: data_mat, iter_num, ratio, k
# Output: avg_emb

for i in range(iter_num):
    # Perform SVD (Eq. 1)
    _, _, Vh = torch.linalg.svd(emb_mat)  
    # Construct the reconstruction matrix (Eq. 2)
    V_r = Vh.T[:, :k]
    recon_mat = V_r @ V_r.T
    # Calculate reconstruction errors (Eq. 3)
    errors = torch.mean((emb_mat - emb_mat @ recon_mat)**2, dim=1)
    # Remove embeddings with largest errors
    m_keep = max(int(emb_mat.shape[0] * ratio), 1)
    _, indices = torch.topk(errors, m_keep, largest=False)
    emb_mat = emb_mat[indices]
    
# Get the average embedding
avg_emb = torch.mean(emb_mat, dim=0, keepdim=True)
\end{lstlisting}
\end{algorithm}

\subsection{\techb}
\label{sec:techb}

While identity-preserving generators excel at maintaining identities in generated images \cite{li2024photomaker, ip-adapter, wang2024instantid}, they exhibit suboptimal performance on text alignment.
This limitation affects their ability to generate desired visual elements and reasonable character layouts, as illustrated in Figure~\ref{fig:motivation_align}(a).
To address this, we design \textit{\techb} (Figure~\ref{fig:pipeline}(b)), utilizing the complementary strengths of general generators and identity-preserving generators.
Specifically, the general generator provides high-quality environmental details and character layouts, while the identity-preserving generator focuses on injecting identities.
This approach involves three processes: noisy image caching, layout extraction, and re-denoising, which are detailed as follows.

\myparagraph{Noisy Image Caching.}
First, we utilize a general generator $G$ (\eg, the base model), which is better in text alignment (Figure~\ref{fig:motivation_align}), to generate template images containing faithful visual semantics for subsequent processes.
For a story image $I_i$, we use its text prompt $T_i$ to generate
\begin{equation}
    I^\prime_i = G(T_i),
\end{equation}
where $I^\prime_i$ is the corresponding template image.
During the sampling of $G$, we cache the noisy images of all  $t$ timesteps, which can be formulated as
\begin{equation}
    \mathbf{z} = \{z_t, z_{t-1}, ..., z_1, z_0\},
\end{equation}
where $z_i \in \mathbb{R}^{l \times l}$ is the noisy image at $i$-th timestep, $l$ is the latent dimension, and $t$ is set to $50$ for a DDIM \cite{song2020denoising} scheduler.
These predicted noisy images implicitly encapsulate the contextual information for generating the template image, serving as the basis for the re-denoising phase.

\begin{figure}[t]
    \centering

    \includegraphics[width=1\linewidth]{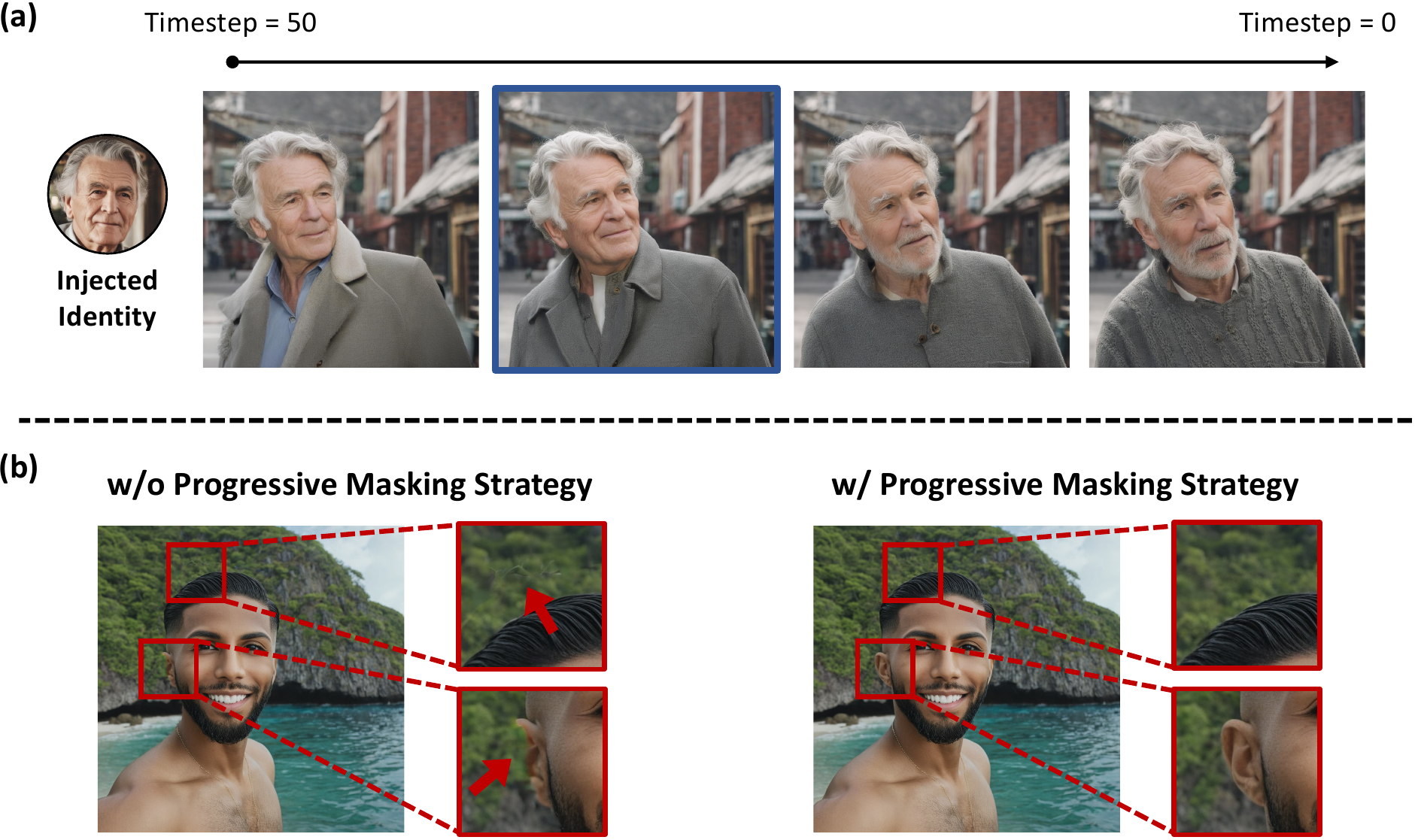}

    \vspace{-3mm}
    \caption
    {
        \textbf{Desgin choices of \techb.}
        (a) We start from a sweet-spot timestep ($t^\prime=40$) to re-denoise, balancing image harmony and identity fidelity (\textit{\textcolor{blue}{blue} frame}).
        (b) We develop a progressive masking strategy to effectively diminish artifacts (\textit{\textcolor{red}{red} frame}).
    }
    \vspace{-3mm}
    
    \label{fig:design_injection}
    
\end{figure}
\begin{table*}[t]

     \caption
     {
         \textbf{Quantitative comparison.}
         The results of automatic metrics demonstrate \method's overall superior performance, especially in \textit{face similarity (Face-Sim)}.
        The best and second-best results are marked in \textbf{bold} and \underline{underlined}.
     }

    \vspace{-2mm}
    
    \centering
    
    \setlength{\tabcolsep}{3mm}  
    \renewcommand{\arraystretch}{1.3}
    
    \resizebox{0.98\linewidth}{!}
    {
\begin{tabular}{lcccccc}
    \specialrule{0.12em}{0pt}{0pt}
    \multirow{2}[4]{*}{Methods} & \multicolumn{2}{c}{Text Alignment} & \multicolumn{2}{c}{Character Consistency} & \multicolumn{2}{c}{Image Quality} \\
    \cmidrule{2-7} 
          & CLIP-T$\uparrow$ (\%) & CLIP-T-C$\uparrow$ (\%) & CLIP-I$\uparrow$ (\%) & Face-Sim$\uparrow$ (\%) & Q-Align-Aes$\uparrow$ & Q-Align-Gen$\uparrow$ \\
    \midrule
    ConsiStory \citep{Consistency} & \textbf{35.5} & 30.1  & 78.2  & 17.1  & 3.75  & 4.71 \\
    StoryDiffusion \citep{StoryDiffusion} & 34.0  & \underline{30.7}  & \underline{85.2}  & \underline{27.1}  & 3.58  & 4.20 \\
    Story-Adapter \citep{mao2024story} & 34.3  & 29.1  & 76.6  & 23.9  & 3.65  & 4.42 \\
    1Prompt1Story \citep{liu2025one} & 34.9  & 29.7  & 79.8  & 23.5  & \underline{4.16}  & \underline{4.81} \\
    \method (Ours) & \underline{35.4}  & \textbf{31.1} & \textbf{85.8} & \textbf{55.5} & \textbf{4.25} & \textbf{4.92} \\
    \specialrule{0.12em}{0pt}{0pt}
\end{tabular}%
    }    

            
    \label{tab:quant_results}
    
\end{table*}
\begin{figure*}[t]
    \centering

    \includegraphics[width=1\linewidth]{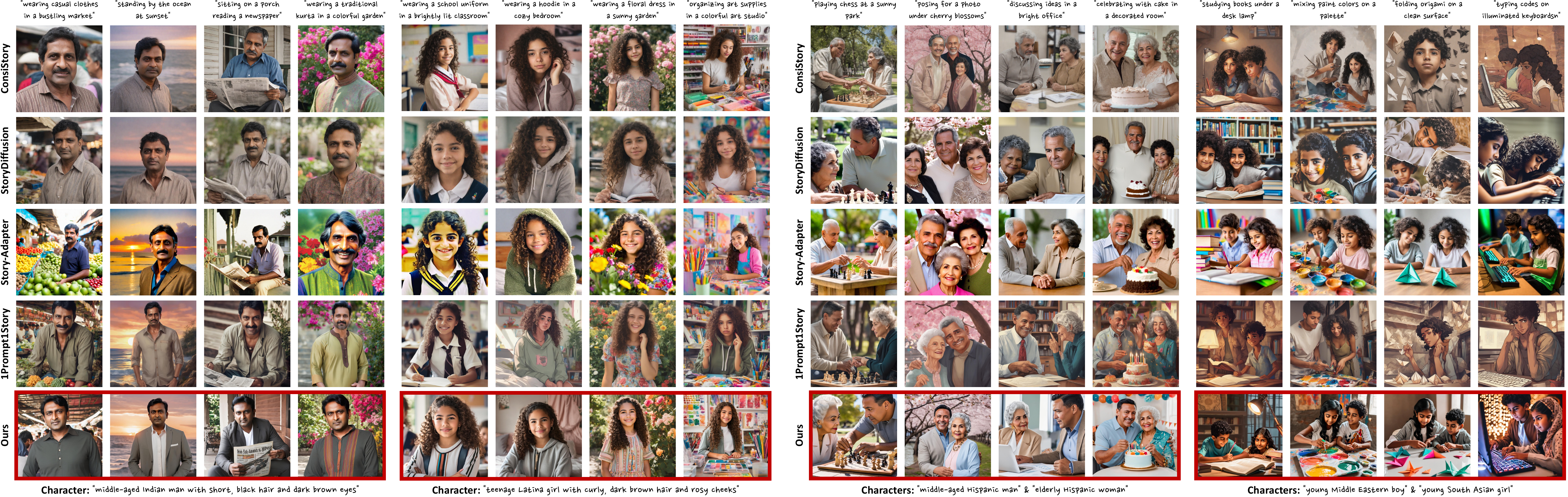}
    \vspace{-6mm}
    \caption
    {
        \textbf{Qualitative comparison.} 
        Compared to other methods, our \method exhibits remarkable performance in handling human-centric scenarios, enabling consistent generation of human characters with only text as input.
        \textit{Zoom in for better view.}
    }
    \vspace{-3mm}
    
    \label{fig:qual_results}
     
\end{figure*}

\myparagraph{Layout Extraction.}
To obtain proper foreground layouts, previous works often rely on user inputs \cite{bar2023multidiffusion,avrahami2023spatext,xie2023boxdiff} or large language models \cite{phung2024grounded,feng2024ranni}.
In contrast, we directly extract the character layout from the template image $I^\prime_i$, as it is generated using the prompt $T_i$, which aligns well with our intended design.
This guarantees that the resultant layout is precise and natural.
Based on character words in $T_i$ such as ``man" and ``woman", we use Grounded-SAM \cite{ren2024grounded} to segment out character-wise masks as
\begin{equation}
    \mathbf{M} = \{{M_1, M_2, ..., M_s}\},
\end{equation}
where the background mask can be set as $M_{\text{bg}} = \complement (M_1 \cup M_2 ... M_s)$.
Using these masks, we can accurately separate each character from the background, allowing for precise identity injection while preserving the original and unbiased environmental details during re-denoising.

\myparagraph{Re-denosing.}
Using the cached noisy images and the character-wise masks, we then inject the identity of characters by performing re-denoising with the identity-preserving generator $G_\text{id}$.
We start from a timestep of $t^\prime=40$, a sweet spot that provides a better balance between image harmony and identity fidelity (Figure~\ref{fig:design_injection}(a)).
For the noisy image $z_{i}$, we use the denoising network $\epsilon_\text{id}$ of $G_\text{id}$ to perform
\begin{equation}
    z_{i-1,j} = \epsilon_\text{id}(c_j, e_j, z_{i}), \quad j = 1, ..., s,
\end{equation}
where $z_{i,j}$ is the noisy image for $j$-th character at the next timestep, which is guided by the identity embedding $e_j$ and the character description $c_j$.
Then, $z_i$ is over-written as
\begin{equation}
    z_{i-1} \leftarrow \text{DS}(M_\text{bg}) \odot z_{i-1} + \sum_{j=1}^s \text{DS}(M_j) \odot z_{i-1,j}, 
\end{equation}
where $\text{DS}(\cdot)$ is a down-sample operation to match the shape of noisy images.
Fixed masks could constrain the scope of re-denoising and make it difficult to adequately reconcile local details, leading to undesired artifacts (Figure~\ref{fig:design_injection}(b)).
Thus, we further develop a progressive masking strategy, updating $M_j$ at each timestep $i$ as
\begin{equation}
    M_j \leftarrow \text{Dilate}(M_j, K_i), \quad j = 1, ..., s,
\end{equation}
where $\text{Dilate}(M_j, K_i)$ denotes dilating the mask $M_j$ with the kernel size $K_i = \frac{i - t^\prime}{t - t^\prime} \cdot K_\text{max}$ and $K_\text{max}$ is the predefined maximum kernel size. This strategy enables re-denoising to progressively refine more local details, effectively eliminating unwanted artifacts (Figure~\ref{fig:design_injection}(b)).

\section{Experiments}
\label{sec:exp}

\subsection{Setups}

\myparagraph{Implementation Details.}
We select PhotoMaker \cite{li2024photomaker} as our identity-preserving generator and SDXL \cite{podell2023sdxl} as the base model. 
For \techa, we set the number of generated character images $m = 64$, the filtering ratio $r = 60\%$, and the iteration number $p = 3$. For \techb, we empirically set the initial timestep $t^\prime = 40$ and the maximum kernel size $M_\text{max}=50$.

\myparagraph{Benchmark and Compared Methods.}
We present an enhanced version of the ConsiStory benchmark \citep{Consistency}, termed \textit{ConsiStory-Human}, which is designed specifically for human-centric scenarios. It features diverse characters and story descriptions, comprising 100 prompt sets, each with 10 text prompts. Using ConsiStory-Human, each method generates 1,000 images for a thorough evaluation. 
We compare our method with state-of-the-art story generation methods, including ConsiStory \citep{Consistency}, StoryDiffusion \citep{StoryDiffusion}, Story-Adapter \citep{mao2024story}, and 1Prompt1Story \citep{liu2025one}. 
More details are provided in Appendix~\ref{app:setup_details}.

\begin{table*}[t]
    \centering
    \begin{minipage}[t]{0.48\linewidth}
    \centering
    \caption
     {
         \textbf{MLLM-as-a-judge evaluation.}
         The proposed \method achieves the highest average across all three metrics, further showcasing its superior performance.
     }
    \vspace{-2mm}
    \setlength{\tabcolsep}{3.4mm}  
    \renewcommand{\arraystretch}{1.3}
    \resizebox{\linewidth}{!}
    {
        \begin{tabular}{lccc}
        \specialrule{0.12em}{0pt}{0pt}
        Methods & Text Align.$\uparrow$ & Char. Consis.$\uparrow$ & Img. Qual.$\uparrow$ \\
        \midrule
    ConsiStory \citep{Consistency} & 81.88 & \underline{78.58} & 82.91 \\
    StoryDiffusion \citep{StoryDiffusion} & 80.74 & 77.52 & 85.47 \\
    Story-Adapter \citep{mao2024story} & 81.08 & 77.00    & 83.07 \\
    1Prompt1Story \citep{liu2025one} & \underline{82.20}  & 78.31 & \underline{86.48} \\
    \method (Ours) & \textbf{84.41} & \textbf{82.83} & \textbf{88.74} \\
        \specialrule{0.12em}{0pt}{0pt}
        \end{tabular}%
    }
    \vspace{-1mm}
    \label{tab:gpt_eval}

    \end{minipage}
    \hfill 
    \begin{minipage}[t]{0.48\linewidth} 
    \centering
    \caption
     {
         \textbf{User study.}
         The selection rates on the three metrics clearly indicate that \method outperforms other methods in terms of human preference.
     }
    \vspace{-2mm}
    \setlength{\tabcolsep}{1.2mm}  
    \renewcommand{\arraystretch}{1.3}
    \resizebox{\linewidth}{!}
    {
        \begin{tabular}{lccc}
        \specialrule{0.12em}{0pt}{0pt}
        Methods & Text Align.$\uparrow$ (\%) & Char. Consis.$\uparrow$ (\%) & Img.Qual.$\uparrow$ (\%) \\
        \midrule
        ConsiStory \citep{Consistency} & 6.5  & 6.5  & 8.5 \\
        StoryDiffusion \citep{StoryDiffusion} & \underline{10.3}  & \underline{11.0}  & \underline{14.3} \\
        Story-Adapter \citep{mao2024story} & 5.0  & 5.2  & 3.8 \\
        1Prompt1Story \citep{liu2025one} & 9.0  & 10.8  & 8.7 \\
        \method (Ours) & \textbf{69.2}  & \textbf{66.5}  & \textbf{64.7} \\
        \specialrule{0.12em}{0pt}{0pt}
        \end{tabular}%
    }
    \vspace{-1mm}
    \label{tab:user_study}
 
    \end{minipage}
\end{table*}

\begin{figure*}[t]
    \centering

    \includegraphics[width=0.98\linewidth]{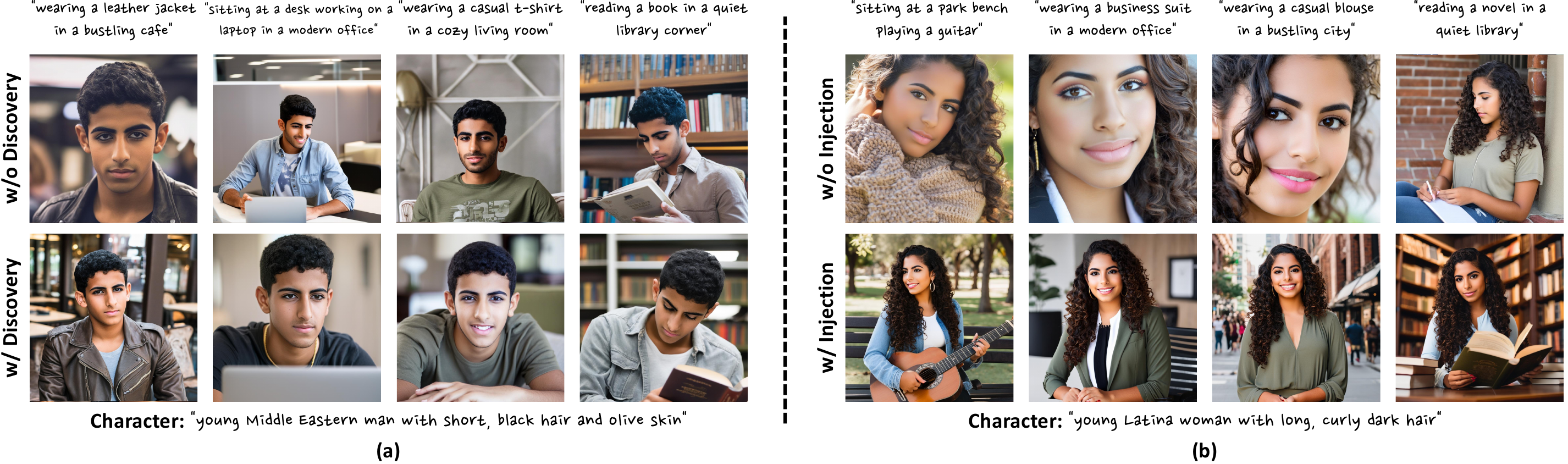}
    
    \vspace{-3mm}
    
    \caption
    {
        \textbf{Ablation of key techniques.}
        We demonstrate the effectiveness of (a) \techa, which improves character consistency, and (b) \techb, which enhances text alignment.
    }
    \vspace{-3mm}
    
    \label{fig:abl_key}
     
\end{figure*}

\myparagraph{Evaluation Metrics.}
We evaluate methods based on the following aspects:
(i) \textit{Text alignment}: We compute the average CLIP \cite{radford2021learning} score between each generated story image and its corresponding text prompt, denoted as CLIP-T. Additionally, we segment characters from each story image and compute the average CLIP score between each character image and its character description, denoted as CLIP-T-C.
(ii) \textit{Character consistency}: We measure the average similarity between segmented character images within each prompt set using CLIP \cite{radford2021learning}, referred to as CLIP-I.
Morover, we compute the average pairwise face similarity with ArcFace~\cite{deng2019arcface}, denoted as Face-Sim.
(iii) \textit{Image quality}: We assess both the aesthetic quality and general quality of the generated story images using Q-Align \cite{wu2023q}, with scores represented as Q-Align-Aes and Q-Align-Gen.

\subsection{Experimental Results}

\myparagraph{Quantitative Comparison.}
We present the quantitative results in Table~\ref{tab:quant_results}.
In terms of text alignment, our method achieves the best CLIP-I-C and performs just slightly below ConsiStory in CLIP-I.
However, the better CLIP-T of ConsiStory comes at the cost of its poor character consistency.
In contrast, our method shows remarkable performance in character consistency, particularly in face similarity.
Specifically, we reach a Face-Sim of 55.5\%, more than double the second-best of 21.1\%.
Moreover, Q-Align indicates that our method can generate top-quality images, excelling in both general and aesthetic aspects.

\myparagraph{Qualitative Comparison.}
The qualitative results are illustrated in Figure~\ref{fig:qual_results} (see more in Appendix~\ref{app:more_qual}).
Our method exhibits superior performance in preserving character consistency, especially for fine-grained facial features, across different images while faithfully following text prompts.
However, other methods struggle with human-centric scenarios, as they rely on attention or prompts to maintain consistency.
Due to its flexibility, our method can also effectively handle multiple characters, whereas other methods cause identity disorder or even blending.
Notably, they also display inconsistencies in style, indicating their instability.

\myparagraph{MLLM-as-a-Judge Evaluation.}
To conduct additional evaluation, we utilize GPT-4o \cite{hurst2024gpt} to score the generated images.
For each case, GPT-4o is provided with the story text alongside the five outputs from other methods and our IdentityStory.
We then instruct GPT-4o to independently assign a score between 0 and 100 to each generated image based on three criteria, including text alignment, character consistency, and image quality.
Finally, we average its ratings for each method and criterion, presenting the results in Table~\ref{tab:gpt_eval}.
As we can observe, IdentityStory achieves the highest averages across all aspects, further demonstrating its superior performance in text alignment, character consistency, and image quality compared to other methods.

\myparagraph{User Study.}
To evaluate human preference, we design a questionnaire showing 20 randomly selected groups of generated images, each paired with the corresponding story text. 
Participants were asked to select the best result in each group in terms of text alignment, character consistency, and image quality. 
In total, we collected 30 responses and the averaged selection rates are reported in Table~\ref{tab:user_study}. 
The results indicate that \method outperforms others across all aspects, showing that its generated images are better aligned with human preference.

\subsection{Ablation Studies}

\myparagraph{Key Techniques.}
We ablate two key techniques of \method to demonstrate their significance.
When removing \techa, we adopt the average identity embedding of the generated character images for the subsequent process, showing that it contributes to superior character consistency (Figure~\ref{fig:abl_key}(a)).
When removing \techb, we use the identity-preserving generator to directly produce final results, verifying that it helps achieve better text alignment (Figure~\ref{fig:abl_key}(b)).

\begin{figure}[t]
    \centering

    \includegraphics[width=0.98\linewidth]{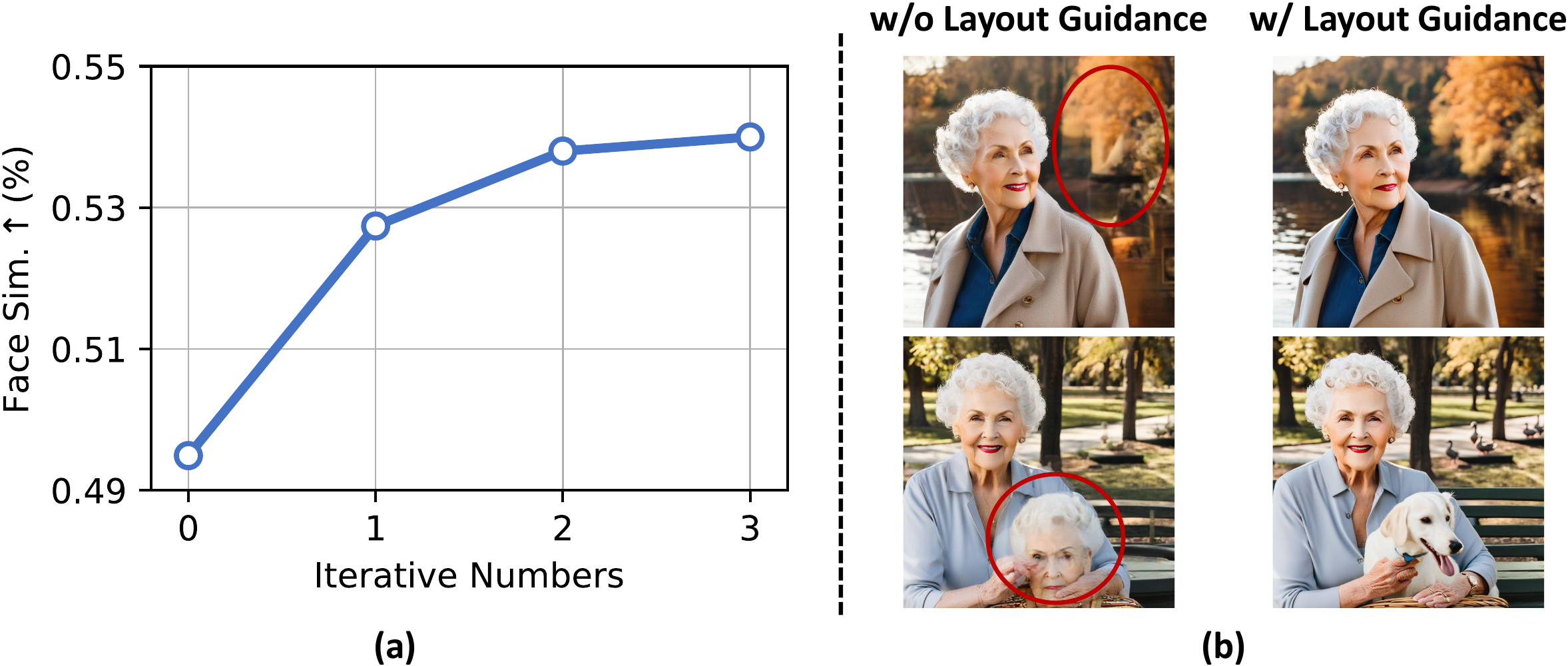}

    \vspace{-2mm}
    \caption
    {
        \textbf{(a) Ablation of the iterative strategy of \techa.}
        Face similarity (Face-Sim) increases steadily with iterative numbers, showing the effectiveness of the iterative strategy.
        \textbf{(b) Ablation of the injection mechanism of \techb.}
        Our layout-guided mechanism can effectively inject identities while maintaining high-quality environmental details.  
    }
    
    \vspace{-1mm}
    
    \label{fig:abl_iter_layout}
    
\end{figure}

\myparagraph{Iterative Strategy of \techa.}
In Figure~\ref{fig:motivation_emb}(b), we have demonstrated that our \techa can achieve better identity embeddings compared to the naive averaging approach.
In Figure~\ref{fig:abl_iter_layout}(a), we follow the same setting and further adjust the iterative numbers of \techa, showing the iterative strategy contributes to a better face similarity result.

\begin{figure}[t]
    \centering

    \includegraphics[width=0.95\linewidth]{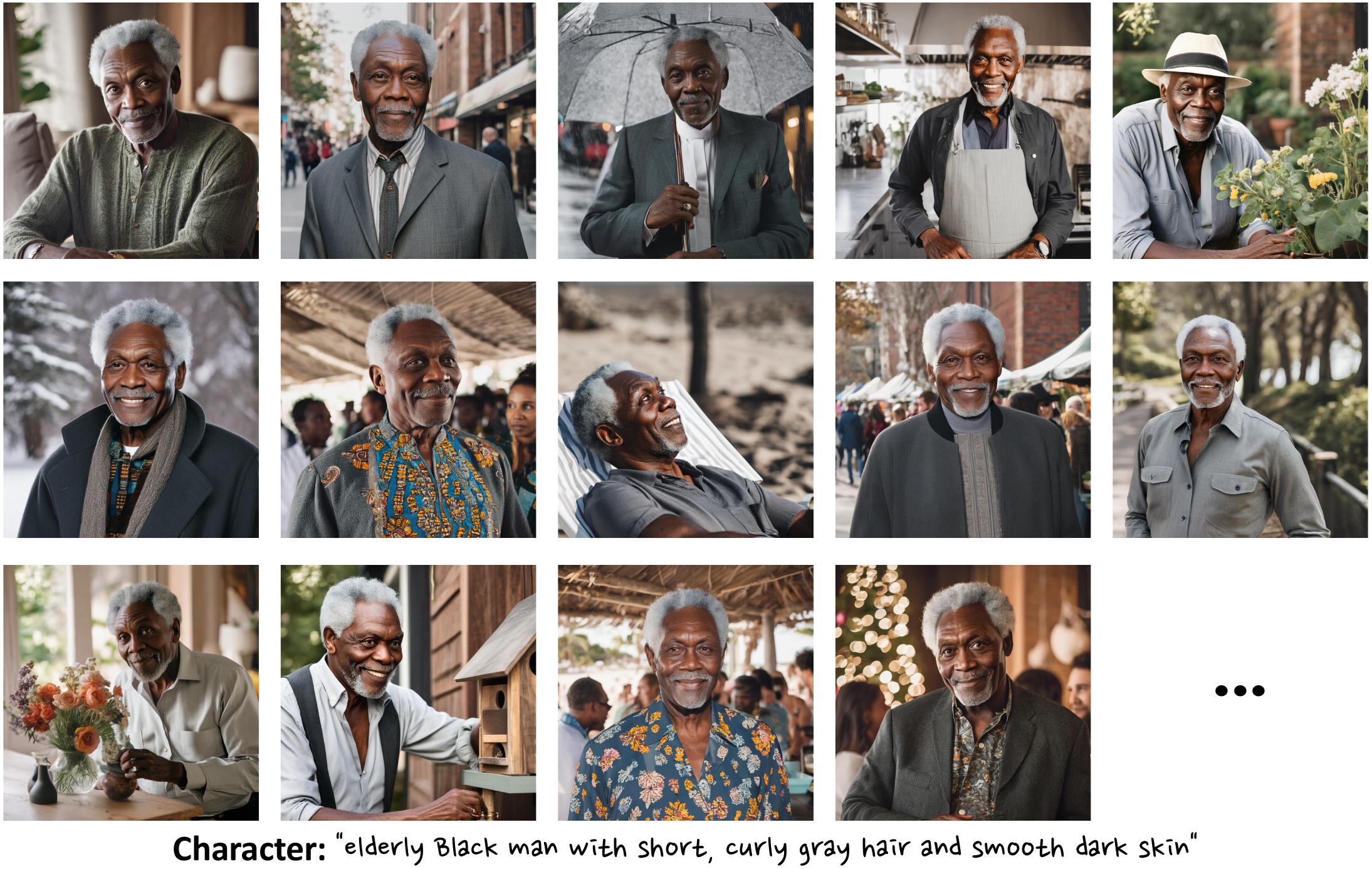}

    \vspace{-2mm}
    \caption
    {
        \textbf{Infinite-length story generation with \method.} 
        Our method can conveniently generate a series of images with consistent characters at any length. 
    }
    \vspace{-3mm}
    
    \label{fig:inf_length}
     
\end{figure}

\myparagraph{Injection Mechanism of \techb.}
We adopt a layout-guided mechanism to precisely inject identities into cached noisy images while preserving environmental details.
To evaluate its effectiveness, we abandon the layout guidance and directly average cached noisy images and the identity generator's output to perform identity injection. 
This paradigm results in disordered visual elements, as it fails to prevent undesired semantics from being mixed into the cached noisy images (Figure~\ref{fig:abl_iter_layout}(b)).

\begin{figure}[t]
    \centering

    \includegraphics[width=0.95\linewidth]{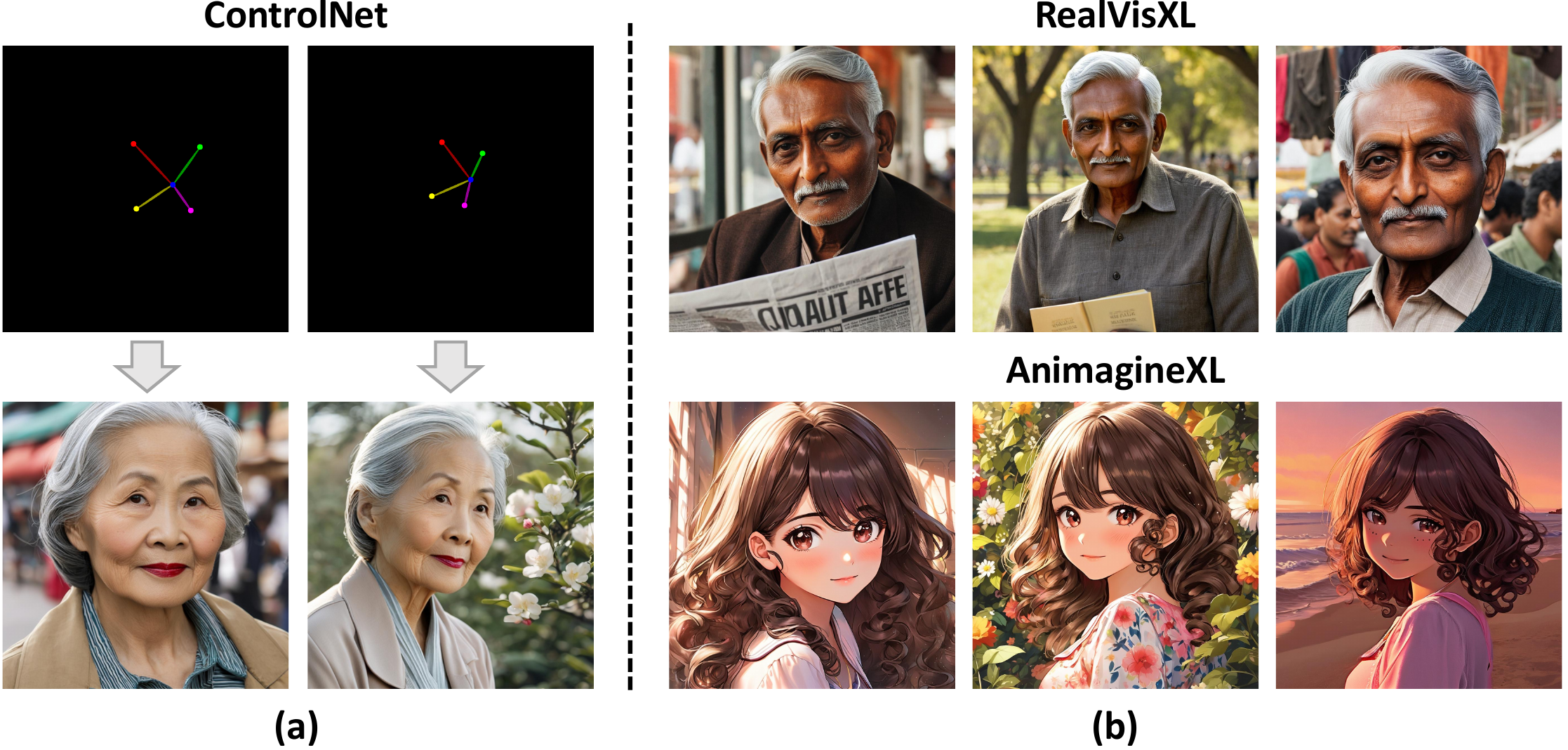}

    \vspace{-2mm}
    \caption
    {
        \textbf{Combining \method with community tools,}
        \method can collaborate with community tools, such as
        (a) ControlNet \cite{zhang2023adding} and 
        (b) stylized base models (RealVisXL \cite{realvisxl} and AnimagineXL \cite{animaginexl}).
    }
    \vspace{-1mm}
    
    \label{fig:app_commu}
    
\end{figure}
\begin{figure}[t]
    \centering

    \includegraphics[width=0.95\linewidth]{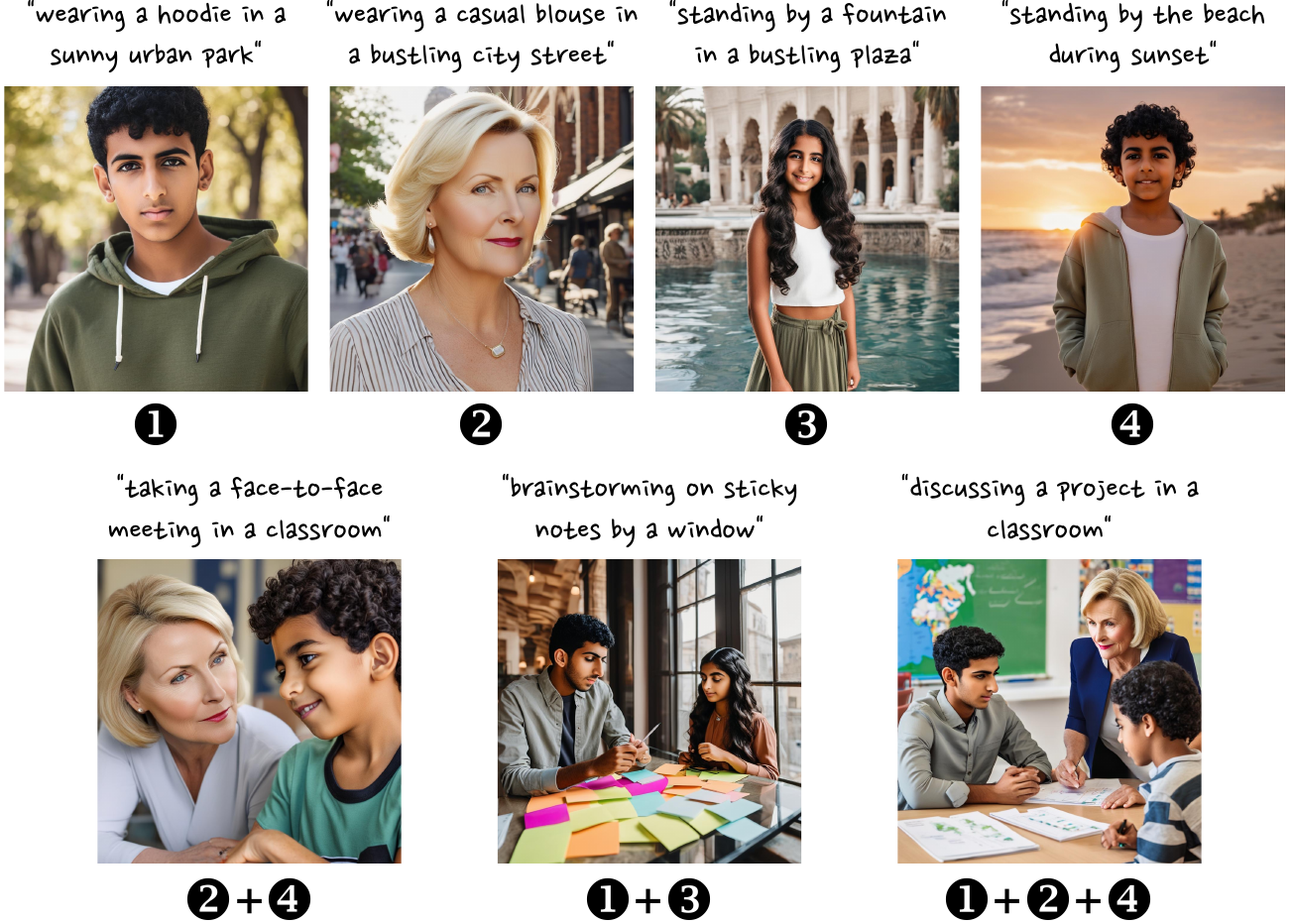}

    \vspace{-2mm}
    
    \caption
    {
        \textbf{Dynamic character composition with \method.} 
        Our method can dynamically combine characters in different story images.
    }
    \vspace{-4mm}
    
    \label{fig:art_char}
     
\end{figure}

\subsection{More Applications}

\myparagraph{Integration with Other Community Tools.}
\method can seamlessly integrate with other community tools to accommodate a wider range of use scenarios (Figure~\ref{fig:app_commu}).
For instance, \method can enable face pose controls with ControlNet \cite{zhang2023adding}, and collaborate with stylized base models \cite{realvisxl, animaginexl} for enhancing the generation of special styles, highlighting its flexibility for diverse user needs.

\myparagraph{Infinite-Length Story Generation.}
One of the key advantages of \method is its ability to easily generate story images of infinite length.
Traditional story generation methods are constrained by memory consumption \cite{Consistency,StoryDiffusion,mao2024story} or text length limitations \cite{liu2025one}, while our method only requires extracting the identity once, after which it can continuously generate images with the same character (Figure~\ref{fig:inf_length}).

\myparagraph{Dynamic Character Composition.}
\method can also dynamically combine characters across different images within a story, as shown in Figure~\ref{fig:art_char}.
This capability allows characters to be flexibly introduced or combined across various scenes, which is highly practical for handling complex narratives without losing coherence or visual harmony.

\section{Conclusion}
\label{sec:concl}
In this work, we introduce the task of \textit{human-centric story generation}, designed to create visual narratives in human-centric scenarios. 
To address its challenges, we develop \textit{\method}, which combines \textit{\techa} to extract identities via iterative filtering and \textit{\techb} to inject identities via re-denoising.
Extensive experiments show that \method sets a new standard for human-centric story generation, highlighting its potential for a wide range of creative applications.
Looking ahead, we plan to extend our framework to include broader image and video generation, further enhancing the coherence, quality, and creativity of generated visual content. Additionally, we aim to explore the integration of multi-modal inputs, such as audio and textual cues, to enrich the storytelling process and create more immersive human-centric narratives.

\section*{Acknowledgments}
This study was supported in part by the InnoHK initiative of the Innovation and Technology Commission of the Hong Kong Special Administrative Region Government via the Hong Kong Centre for Logistics Robotics, by the Faculty Initiatives Research of Monash University (Contract No. 2901912), by the NVIDIA Academic Hardware Grant Program, and by the Research Start-up Fund for Prof. Xiaowei Hu at the Guangzhou International Campus, South China University of Technology (Grant No. K3250310).

\bibliography{aaai2026}

@String(TOG= {ACM Trans. Graph.})

@String(TOG   = {ACM TOG})

@inproceedings{rombach2022high,
  title={High-resolution image synthesis with latent diffusion models},
  author={Rombach, Robin and Blattmann, Andreas and Lorenz, Dominik and Esser, Patrick and Ommer, Bj{\"o}rn},
  booktitle={Proceedings of the IEEE/CVF conference on computer vision and pattern recognition},
  pages={10684--10695},
  year={2022}
}

@article{podell2023sdxl,
  title={Sdxl: Improving latent diffusion models for high-resolution image synthesis},
  author={Podell, Dustin and English, Zion and Lacey, Kyle and Blattmann, Andreas and Dockhorn, Tim and M{\"u}ller, Jonas and Penna, Joe and Rombach, Robin},
  journal={arXiv preprint arXiv:2307.01952},
  year={2023}
}

@article{klimmt2012forecasting,
  title={Forecasting the Experience of Future Entertainment Technology: “Interactive Storytelling” and Media Enjoyment},
  author={Klimmt, Christoph and Roth, Christian and Vermeulen, Ivar and Vorderer, Peter and Roth, Franziska Susanne},
  journal={Games and Culture},
  volume={7},
  number={3},
  pages={187--208},
  year={2012},
  publisher={Sage Publications Sage CA: Los Angeles, CA}
}

@article{carter1993place,
  title={The place of story in the study of teaching and teacher education},
  author={Carter, Kathy},
  journal={Educational researcher},
  volume={22},
  number={1},
  pages={5--18},
  year={1993},
  publisher={Sage Publications Sage CA: Thousand Oaks, CA}
}

@book{hart2013art,
  title={The Art of the Storyboard: A filmmaker's introduction},
  author={Hart, John},
  year={2013},
  publisher={Routledge}
}

@book{halligan2013movie,
  title={Movie Storyboards: The art of visualizing screenplays},
  author={Halligan, Fionnuala},
  year={2013},
  publisher={Chronicle Books}
}

@incollection{escalas2003advertising,
  title={Advertising narratives: what are they and how do they work?},
  author={Escalas, Jennifer Edson},
  booktitle={Representing consumers},
  pages={283--305},
  year={2003},
  publisher={Routledge}
}

@article{megehee2010creating,
  title={Creating visual narrative art for decoding stories that consumers and brands tell},
  author={Megehee, Carol M and Woodside, Arch G},
  journal={Psychology \& Marketing},
  volume={27},
  number={6},
  pages={603--622},
  year={2010},
  publisher={Wiley Online Library}
}

@article{cetinic2022understanding,
  title={Understanding and creating art with AI: Review and outlook},
  author={Cetinic, Eva and She, James},
  journal={ACM Transactions on Multimedia Computing, Communications, and Applications (TOMM)},
  volume={18},
  number={2},
  pages={1--22},
  year={2022},
  publisher={ACM New York, NY}
}

@article{zhou2024magictailor,
  title={{MagicTailor}: Component-Controllable Personalization in Text-to-Image Diffusion Models},
  author={Zhou, Donghao and Huang, Jiancheng and Bai, Jinbin and Wang, Jiaze and Chen, Hao and Chen, Guangyong and Hu, Xiaowei and Heng, Pheng-Ann},
  journal={arXiv preprint arXiv:2410.13370},
  year={2024}
}

@inproceedings{zhang2023adding,
  title={Adding conditional control to text-to-image diffusion models},
  author={Zhang, Lvmin and Rao, Anyi and Agrawala, Maneesh},
  booktitle={Proceedings of the IEEE/CVF International Conference on Computer Vision},
  year={2023}
}

@article{ip-adapter,
  title={Ip-adapter: Text compatible image prompt adapter for text-to-image diffusion models},
  author={Ye, Hu and Zhang, Jun and Liu, Sibo and Han, Xiao and Yang, Wei},
  journal={arXiv preprint arXiv:2308.06721},
  year={2023}
}

@article{Consistency,
  title={Training-free consistent text-to-image generation},
  author={Tewel, Yoad and Kaduri, Omri and Gal, Rinon and Kasten, Yoni and Wolf, Lior and Chechik, Gal and Atzmon, Yuval},
  journal={ACM Transactions on Graphics (TOG)},
  year={2024},
}

@article{StoryDiffusion,
  title={Storydiffusion: Consistent self-attention for long-range image and video generation},
  author={Zhou, Yupeng and Zhou, Daquan and Cheng, Ming-Ming and Feng, Jiashi and Hou, Qibin},
  journal={Advances in Neural Information Processing Systems},
  volume={37},
  pages={110315--110340},
  year={2024}
}

@article{gal2022image,
  title={An image is worth one word: Personalizing text-to-image generation using textual inversion},
  author={Gal, Rinon and Alaluf, Yuval and Atzmon, Yuval and Patashnik, Or and Bermano, Amit H and Chechik, Gal and Cohen-Or, Daniel},
  journal={arXiv preprint arXiv:2208.01618},
  year={2022}
}

@inproceedings{ruiz2023dreambooth,
  title={Dreambooth: Fine tuning text-to-image diffusion models for subject-driven generation},
  author={Ruiz, Nataniel and Li, Yuanzhen and Jampani, Varun and Pritch, Yael and Rubinstein, Michael and Aberman, Kfir},
  booktitle={Proceedings of the IEEE/CVF conference on computer vision and pattern recognition},
  year={2023}
}

@inproceedings{li2019storygan,
  title={Storygan: A sequential conditional gan for story visualization},
  author={Li, Yitong and Gan, Zhe and Shen, Yelong and Liu, Jingjing and Cheng, Yu and Wu, Yuexin and Carin, Lawrence and Carlson, David and Gao, Jianfeng},
  booktitle={Proceedings of the IEEE/CVF conference on computer vision and pattern recognition},
  year={2019}
}

@article{gong2023talecrafter,
  title={Talecrafter: Interactive story visualization with multiple characters},
  author={Gong, Yuan and Pang, Youxin and Cun, Xiaodong and Xia, Menghan and He, Yingqing and Chen, Haoxin and Wang, Longyue and Zhang, Yong and Wang, Xintao and Shan, Ying and others},
  journal={arXiv preprint arXiv:2305.18247},
  year={2023}
}

@article{shen2023large,
  title={Large Language Models as Consistent Story Visualizers},
  author={Shen, Xiaoqian and Elhoseiny, Mohamed},
  journal={arXiv preprint arXiv:2312.02252},
  year={2023}
}

@article{yang2024seed,
  title={Seed-story: Multimodal long story generation with large language model},
  author={Yang, Shuai and Ge, Yuying and Li, Yang and Chen, Yukang and Ge, Yixiao and Shan, Ying and Chen, Yingcong},
  journal={arXiv preprint arXiv:2407.08683},
  year={2024}
}

@misc{FLUX,
    author={{Black Forest Labs}},
    title={FLUX},
    year={2024},
    howpublished={\url{https://github.com/black-forest-labs/flux}},
}

@article{mao2024story,
  title={Story-Adapter: A Training-free Iterative Framework for Long Story Visualization},
  author={Mao, Jiawei and Huang, Xiaoke and Xie, Yunfei and Chang, Yuanqi and Hui, Mude and Xu, Bingjie and Zhou, Yuyin},
  journal={arXiv preprint arXiv:2410.06244},
  year={2024}
}

@inproceedings{liu2024intelligent,
  title={Intelligent Grimm-Open-ended Visual Storytelling via Latent Diffusion Models},
  author={Liu, Chang and Wu, Haoning and Zhong, Yujie and Zhang, Xiaoyun and Wang, Yanfeng and Xie, Weidi},
  booktitle={Proceedings of the IEEE/CVF Conference on Computer Vision and Pattern Recognition},
  pages={6190--6200},
  year={2024}
}

@inproceedings{li2024photomaker,
  title={Photomaker: Customizing realistic human photos via stacked id embedding},
  author={Li, Zhen and Cao, Mingdeng and Wang, Xintao and Qi, Zhongang and Cheng, Ming-Ming and Shan, Ying},
  booktitle={Proceedings of the IEEE/CVF Conference on Computer Vision and Pattern Recognition},
  pages={8640--8650},
  year={2024}
}

@inproceedings{maharana2022storydall,
  title={Storydall-e: Adapting pretrained text-to-image transformers for story continuation},
  author={Maharana, Adyasha and Hannan, Darryl and Bansal, Mohit},
  booktitle={European Conference on Computer Vision},
  pages={70--87},
  year={2022},
  organization={Springer}
}

@article{li2020improved,
  title={Improved-storygan for sequential images visualization},
  author={Li, Chunye and Kong, Liya and Zhou, Zhiping},
  journal={Journal of Visual Communication and Image Representation},
  volume={73},
  pages={102956},
  year={2020},
  publisher={Elsevier}
}

@inproceedings{liu2025one,
  title={One-Prompt-One-Story: Free-Lunch Consistent Text-to-Image Generation Using a Single Prompt},
  author={Liu, Tao and Wang, Kai and Li, Senmao and van de Weijer, Joost and Khan, Fahad Shahbaz and Yang, Shiqi and Wang, Yaxing and Yang, Jian and Cheng, Ming-Ming},
  booktitle={The Thirteenth International Conference on Learning Representations},
    year={2025}
}

@article{wang2021deep,
  title={Deep face recognition: A survey},
  author={Wang, Mei and Deng, Weihong},
  journal={Neurocomputing},
  volume={429},
  pages={215--244},
  year={2021},
  publisher={Elsevier}
}

@inproceedings{zhou2021face,
  title={Face forensics in the wild},
  author={Zhou, Tianfei and Wang, Wenguan and Liang, Zhiyuan and Shen, Jianbing},
  booktitle={Proceedings of the IEEE/CVF conference on computer vision and pattern recognition},
  pages={5778--5788},
  year={2021}
}

@article{kammoun2022generative,
  title={Generative adversarial networks for face generation: A survey},
  author={Kammoun, Amina and Slama, Rim and Tabia, Hedi and Ouni, Tarek and Abid, Mohmed},
  journal={ACM Computing Surveys},
  volume={55},
  number={5},
  pages={1--37},
  year={2022},
  publisher={ACM New York, NY}
}

@article{he2024disenvisioner,
  title={DisEnvisioner: Disentangled and Enriched Visual Prompt for Customized Image Generation},
  author={He, Jing and Li, Haodong and Hu, Yongzhe and Shen, Guibao and Cai, Yingjie and Qiu, Weichao and Chen, Ying-Cong},
  journal={arXiv preprint arXiv:2410.02067},
  year={2024}
}

@inproceedings{wei2023elite,
  title={Elite: Encoding visual concepts into textual embeddings for customized text-to-image generation},
  author={Wei, Yuxiang and Zhang, Yabo and Ji, Zhilong and Bai, Jinfeng and Zhang, Lei and Zuo, Wangmeng},
  booktitle={Proceedings of the IEEE/CVF International Conference on Computer Vision},
  pages={15943--15953},
  year={2023}
}

@article{xiao2024fastcomposer,
  title={Fastcomposer: Tuning-free multi-subject image generation with localized attention},
  author={Xiao, Guangxuan and Yin, Tianwei and Freeman, William T and Durand, Fr{\'e}do and Han, Song},
  journal={International Journal of Computer Vision},
  pages={1--20},
  year={2024},
  publisher={Springer}
}

@inproceedings{kumari2023multi,
  title={Multi-concept customization of text-to-image diffusion},
  author={Kumari, Nupur and Zhang, Bingliang and Zhang, Richard and Shechtman, Eli and Zhu, Jun-Yan},
  booktitle={Proceedings of the IEEE/CVF Conference on Computer Vision and Pattern Recognition},
  pages={1931--1941},
  year={2023}
}

@article{wang2024instantid,
  title={Instantid: Zero-shot identity-preserving generation in seconds},
  author={Wang, Qixun and Bai, Xu and Wang, Haofan and Qin, Zekui and Chen, Anthony and Li, Huaxia and Tang, Xu and Hu, Yao},
  journal={arXiv preprint arXiv:2401.07519},
  year={2024}
}

@article{rout2024rb,
  title={RB-Modulation: Training-Free Personalization of Diffusion Models using Stochastic Optimal Control},
  author={Rout, Litu and Chen, Yujia and Ruiz, Nataniel and Kumar, Abhishek and Caramanis, Constantine and Shakkottai, Sanjay and Chu, Wen-Sheng},
  journal={arXiv preprint arXiv:2405.17401},
  year={2024}
}

@article{jiang2024mc,
  title={MC $^2$: Multi-concept Guidance for Customized Multi-concept Generation},
  author={Jiang, Jiaxiu and Zhang, Yabo and Feng, Kailai and Wu, Xiaohe and Li, Wenbo and Pei, Renjing and Li, Fan and Zuo, Wangmeng},
  journal={arXiv preprint arXiv:2404.05268},
  year={2024}
}

@article{tan2024ominicontrol,
  title={Ominicontrol: Minimal and universal control for diffusion transformer},
  author={Tan, Zhenxiong and Liu, Songhua and Yang, Xingyi and Xue, Qiaochu and Wang, Xinchao},
  journal={arXiv preprint arXiv:2411.15098},
  volume={3},
  year={2024}
}

@article{cai2024diffusion,
  title={Diffusion self-distillation for zero-shot customized image generation},
  author={Cai, Shengqu and Chan, Eric and Zhang, Yunzhi and Guibas, Leonidas and Wu, Jiajun and Wetzstein, Gordon},
  journal={arXiv preprint arXiv:2411.18616},
  year={2024}
}

@article{wang2024stableidentity,
  title={Stableidentity: Inserting anybody into anywhere at first sight},
  author={Wang, Qinghe and Jia, Xu and Li, Xiaomin and Li, Taiqing and Ma, Liqian and Zhuge, Yunzhi and Lu, Huchuan},
  journal={arXiv preprint arXiv:2401.15975},
  year={2024}
}

@article{guo2025pulid,
  title={Pulid: Pure and lightning id customization via contrastive alignment},
  author={Guo, Zinan and Wu, Yanze and Zhuowei, Chen and Zhang, Peng and He, Qian and others},
  journal={Advances in Neural Information Processing Systems},
  volume={37},
  pages={36777--36804},
  year={2025}
}

@article{wei2025personalized,
  title={Personalized Image Generation with Deep Generative Models: A Decade Survey},
  author={Wei, Yuxiang and Zheng, Yiheng and Zhang, Yabo and Liu, Ming and Ji, Zhilong and Zhang, Lei and Zuo, Wangmeng},
  journal={arXiv preprint arXiv:2502.13081},
  year={2025}
}

@article{ren2024grounded,
  title={Grounded sam: Assembling open-world models for diverse visual tasks},
  author={Ren, Tianhe and Liu, Shilong and Zeng, Ailing and Lin, Jing and Li, Kunchang and Cao, He and Chen, Jiayu and Huang, Xinyu and Chen, Yukang and Yan, Feng and others},
  journal={arXiv preprint arXiv:2401.14159},
  year={2024}
}

@article{wu2023q,
  title={Q-align: Teaching lmms for visual scoring via discrete text-defined levels},
  author={Wu, Haoning and Zhang, Zicheng and Zhang, Weixia and Chen, Chaofeng and Liao, Liang and Li, Chunyi and Gao, Yixuan and Wang, Annan and Zhang, Erli and Sun, Wenxiu and others},
  journal={arXiv preprint arXiv:2312.17090},
  year={2023}
}

@inproceedings{radford2021learning,
  title={Learning transferable visual models from natural language supervision},
  author={Radford, Alec and Kim, Jong Wook and Hallacy, Chris and Ramesh, Aditya and Goh, Gabriel and Agarwal, Sandhini and Sastry, Girish and Askell, Amanda and Mishkin, Pamela and Clark, Jack and others},
  booktitle={International conference on machine learning},
  pages={8748--8763},
  year={2021},
  organization={PmLR}
}

@article{hurst2024gpt,
  title={Gpt-4o system card},
  author={Hurst, Aaron and Lerer, Adam and Goucher, Adam P and Perelman, Adam and Ramesh, Aditya and Clark, Aidan and Ostrow, AJ and Welihinda, Akila and Hayes, Alan and Radford, Alec and others},
  journal={arXiv preprint arXiv:2410.21276},
  year={2024}
}

@article{song2020denoising,
  title={Denoising diffusion implicit models},
  author={Song, Jiaming and Meng, Chenlin and Ermon, Stefano},
  journal={arXiv preprint arXiv:2010.02502},
  year={2020}
}

@inproceedings{deng2019arcface,
  title={Arcface: Additive angular margin loss for deep face recognition},
  author={Deng, Jiankang and Guo, Jia and Xue, Niannan and Zafeiriou, Stefanos},
  booktitle={Proceedings of the IEEE/CVF conference on computer vision and pattern recognition},
  pages={4690--4699},
  year={2019}
}

@article{van2008visualizing,
  title={Visualizing data using t-SNE.},
  author={Van der Maaten, Laurens and Hinton, Geoffrey},
  journal={Journal of machine learning research},
  volume={9},
  number={11},
  year={2008}
}

@article{chen2023pixart,
  title={Pixart-$\alpha$: Fast training of diffusion transformer for photorealistic text-to-image synthesis},
  author={Chen, Junsong and Yu, Jincheng and Ge, Chongjian and Yao, Lewei and Xie, Enze and Wu, Yue and Wang, Zhongdao and Kwok, James and Luo, Ping and Lu, Huchuan and others},
  journal={arXiv preprint arXiv:2310.00426},
  year={2023}
}

@inproceedings{gu2014weighted,
  title={Weighted nuclear norm minimization with application to image denoising},
  author={Gu, Shuhang and Zhang, Lei and Zuo, Wangmeng and Feng, Xiangchu},
  booktitle={Proceedings of the IEEE conference on computer vision and pattern recognition},
  pages={2862--2869},
  year={2014}
}

@article{bar2023multidiffusion,
  title={Multidiffusion: Fusing diffusion paths for controlled image generation},
  author={Bar-Tal, Omer and Yariv, Lior and Lipman, Yaron and Dekel, Tali},
  year={2023}
}

@inproceedings{phung2024grounded,
  title={Grounded text-to-image synthesis with attention refocusing},
  author={Phung, Quynh and Ge, Songwei and Huang, Jia-Bin},
  booktitle={Proceedings of the IEEE/CVF Conference on Computer Vision and Pattern Recognition},
  pages={7932--7942},
  year={2024}
}

@inproceedings{avrahami2023spatext,
  title={Spatext: Spatio-textual representation for controllable image generation},
  author={Avrahami, Omri and Hayes, Thomas and Gafni, Oran and Gupta, Sonal and Taigman, Yaniv and Parikh, Devi and Lischinski, Dani and Fried, Ohad and Yin, Xi},
  booktitle={Proceedings of the IEEE/CVF Conference on Computer Vision and Pattern Recognition},
  pages={18370--18380},
  year={2023}
}

@inproceedings{xie2023boxdiff,
  title={Boxdiff: Text-to-image synthesis with training-free box-constrained diffusion},
  author={Xie, Jinheng and Li, Yuexiang and Huang, Yawen and Liu, Haozhe and Zhang, Wentian and Zheng, Yefeng and Shou, Mike Zheng},
  booktitle={Proceedings of the IEEE/CVF International Conference on Computer Vision},
  pages={7452--7461},
  year={2023}
}

@inproceedings{feng2024ranni,
  title={Ranni: Taming text-to-image diffusion for accurate instruction following},
  author={Feng, Yutong and Gong, Biao and Chen, Di and Shen, Yujun and Liu, Yu and Zhou, Jingren},
  booktitle={Proceedings of the IEEE/CVF Conference on Computer Vision and Pattern Recognition},
  pages={4744--4753},
  year={2024}
}

@misc{animaginexl,
  title={Animagine XL 3.0},
  author={{Cagliostro Research Lab}},
  year={2024},
  url={https://huggingface.co/cagliostrolab/animagine-xl-3.0}
}

@misc{realvisxl,
  title={RealVisXL V4.0},
  author={{SG161222}},
  year={2024},
  url={https://huggingface.co/SG161222}
}

@inproceedings{breunig2000lof,
  title={LOF: identifying density-based local outliers},
  author={Breunig, Markus M and Kriegel, Hans-Peter and Ng, Raymond T and Sander, J{\"o}rg},
  booktitle={Proceedings of the 2000 ACM SIGMOD international conference on Management of data},
  pages={93--104},
  year={2000}
}

@inproceedings{ester1996density,
  title={A density-based algorithm for discovering clusters in large spatial databases with noise},
  author={Ester, Martin and Kriegel, Hans-Peter and Sander, J{\"o}rg and Xu, Xiaowei and others},
  booktitle={kdd},
  volume={96},
  number={34},
  pages={226--231},
  year={1996}
}

@article{mahalanobis2018generalized,
  title={On the generalized distance in statistics},
  author={Mahalanobis, Prasanta Chandra},
  journal={Sankhy{\=a}: The Indian Journal of Statistics, Series A (2008-)},
  volume={80},
  pages={S1--S7},
  year={2018},
  publisher={JSTOR}
}

@inproceedings{kong2024omg,
  title={Omg: Occlusion-friendly personalized multi-concept generation in diffusion models},
  author={Kong, Zhe and Zhang, Yong and Yang, Tianyu and Wang, Tao and Zhang, Kaihao and Wu, Bizhu and Chen, Guanying and Liu, Wei and Luo, Wenhan},
  booktitle={European Conference on Computer Vision},
  pages={253--270},
  year={2024},
  organization={Springer}
}

\clearpage

\appendix

\setcounter{figure}{11}

\setcounter{secnumdepth}{2}

\section{More Details of Setups}
\label{app:setup_details}

\subsection{Benchmark Construction}
To systemically investigate method performance on human-centric story generation, we construct \textit{ConsiStory-Human}, a new benchmark improved from the original ConsiStory benchmark \cite{Consistency}.
Specifically, we iteratively instruct GPT-4o to generate the necessary content for generating a human-centric narrative, which is organized as a JSON file containing the following metadata: 
\begin{itemize}[itemsep=0.5em, parsep=0em, left=1em]
    \item
    \textbf{Character category:}
    a list of single words summarizing the human characters of the story, which should be contained in the character description.
    \item 
    \textbf{Character description:}
    a list of detailed appearance descriptions for human characters. We ask the model to produce the extended description based on the character category. The description should only include age, skin color, gender, hairstyle, and facial features. It should avoid depicting expressions, clothing, or any other elements unrelated to the character's identity. The focus is on appealing and natural appearances, avoiding odd or unrealistic combinations, with an emphasis on diversity rather than a specific demographic. The description is limited to 15 words or fewer.
    \item 
    \textbf{Story setting:}
    a list of specific scene or action settings for formulating text prompts. Each setting includes minor story changes and focuses on plots about daily life, emphasizing realistic and ordinary activities or environments. There are no references to artistic, fantasy, surreal, or fictional elements. We also require the model to keep the language simple and clear.
    \item 
    \textbf{Text prompt:} a list of sentences used to guide the generation of T2I models to yield story images. The text prompts are composed directly with story settings and character descriptions.     
\end{itemize}
In total, we generated 100 special stories, each of which contains 10 text prompts,  spanning diverse human characters, scenes, and actions. 
Based on the benchmark, each method generates 1,000 images for a comprehensive evaluation. 
We implement other methods by following their official configurations with the same base model.

\subsection{User Study Design}
In the user study, we designed a questionnaire featuring 20 groups of images generated by our method and other methods, where each group corresponds to a special story.
For each group, the associated text prompts were also provided.
The results from our method and other methods were displayed on the same page.
Clear evaluation criteria were established, focusing on three aspects: text alignment, character consistency, and image quality. Users were asked to select the best result in each group by answering questions based on these three aspects. To ensure fairness, all method names were anonymized.
We hide all the method names to ensure fairness. 
Finally, a total of 30 participants were involved for an evaluation of human preferences.

\begin{figure}[t]
    \centering

    \includegraphics[width=0.5\linewidth]{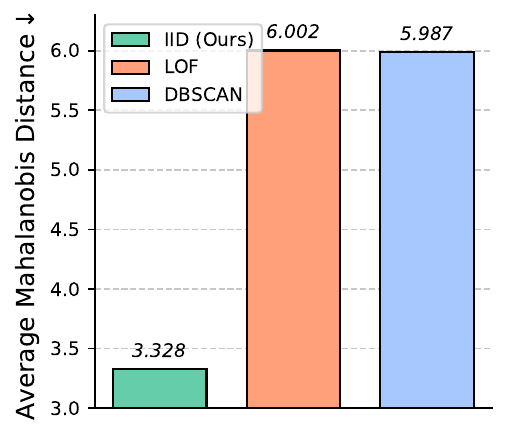}

    \vspace{-3mm}
    \caption
    {
        \textbf{Numerical analysis results of \techa.} 
        \techa achieves the lowest average Mahalanobis distance, indicating its effectiveness.
    }
    \vspace{-3mm}
    
    \label{fig:num_anal}
     
\end{figure}

\section{Additional Results of \method}

\subsection{Numerical Analysis of \techa}
To further verify the effectiveness of \techa, we compare it with two classical outlier exclusion algorithms, Local Outlier Factor (LOF)~\cite{breunig2000lof} and Density-Based Spatial Clustering of Applications with Noise (DBSCAN)~\cite{ester1996density}, in handling the filtering of identity embeddings.
To quantify the compactness of the retained identity embeddings, we measure the average Mahalanobis distance~\cite{mahalanobis2018generalized} between each embedding and the mean embedding of the group.
The results based on 100 groups are reported in Figure~\ref{fig:num_anal}.
As we can observe, \techa achieves the most compact cluster in the identity space, demonstrating the advantage of its specialized design in this scenario. 

\subsection{Seed Variation}
We further evaluate whether our \techa can obtain different identity embeddings with the same character description via seed variation.
Specifically, we use the same character description to run our \techa with different seeds, and generate the character images using the text prompt of ``a portrait of [character]". 
As shown in Figure~\ref{fig:var_style}(a), our method can avoid collapsing into an ``average face" and is able to extract different identities for generating diverse characters.

\subsection{Stylized Story Generation}
\begin{figure*}[t]
    \centering

    \includegraphics[width=1\linewidth]{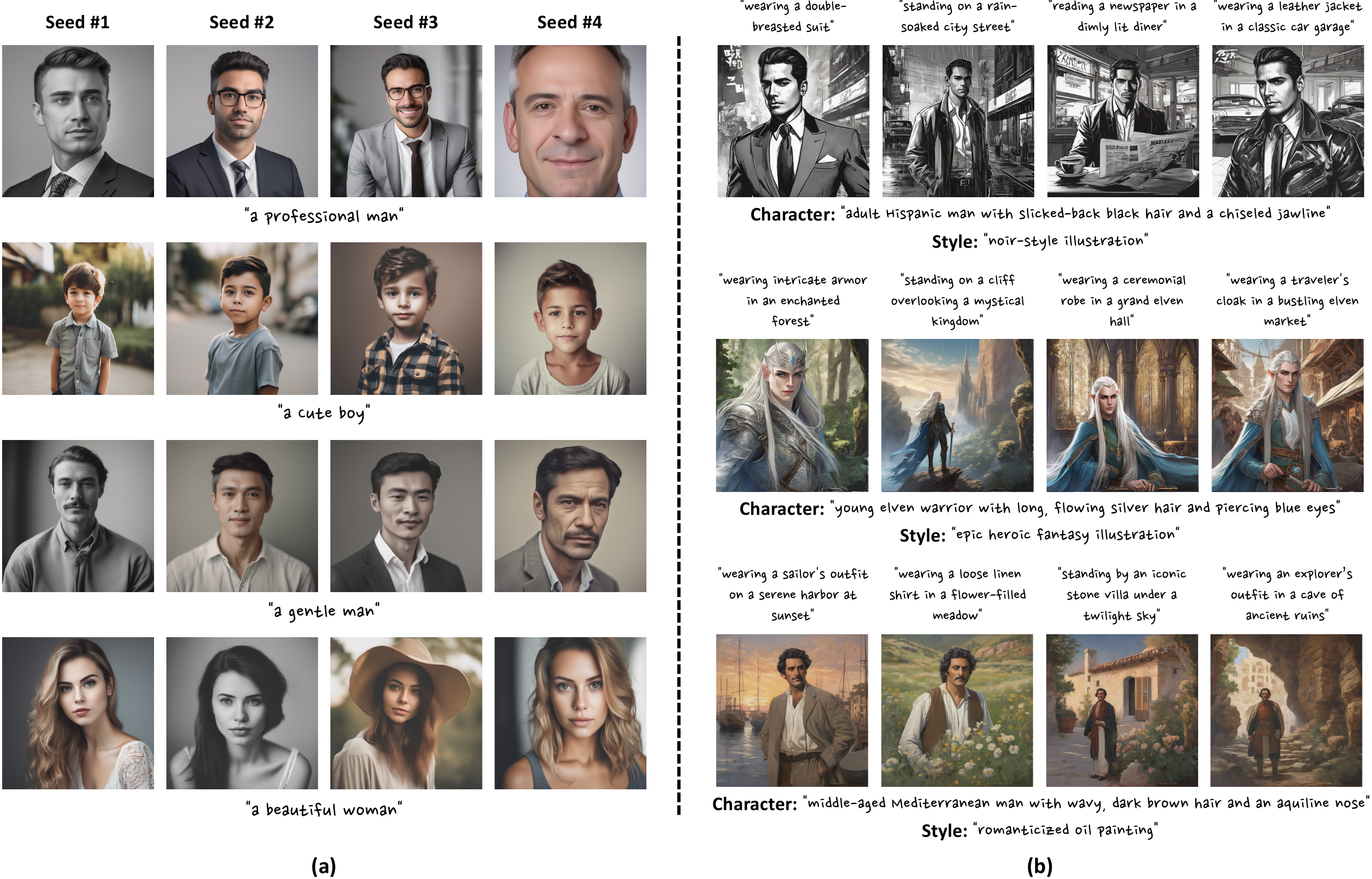}
    
    \vspace{-2mm}
    \caption
    {
        \textbf{(a) Seed variation for generating diverse characters.}
        \techa can extract different identities for a character description via different seeds.
        \textbf{(b) Stylized story generation with \method.}
        \method is also capable of generating diverse visual narratives across different artistic styles while preserving the character's identity across images. 
    }
    \vspace{-5mm}
    
    \label{fig:var_style}
     
\end{figure*}

\method can also generate diverse visual narratives in a wide range of artistic styles, showcasing its versatility and creativity.
By simply appending a style description to the text prompt, \method enables the creation of storytelling that caters to various aesthetic preferences and thematic needs, while effectively preserving the identity across different stylized story images, as illustrated in Figure~\ref{fig:var_style}(b).
 
\section{More Qualitative Results}
\label{app:more_qual}
\begin{figure*}[t]
    \centering

    \includegraphics[width=0.95\linewidth]{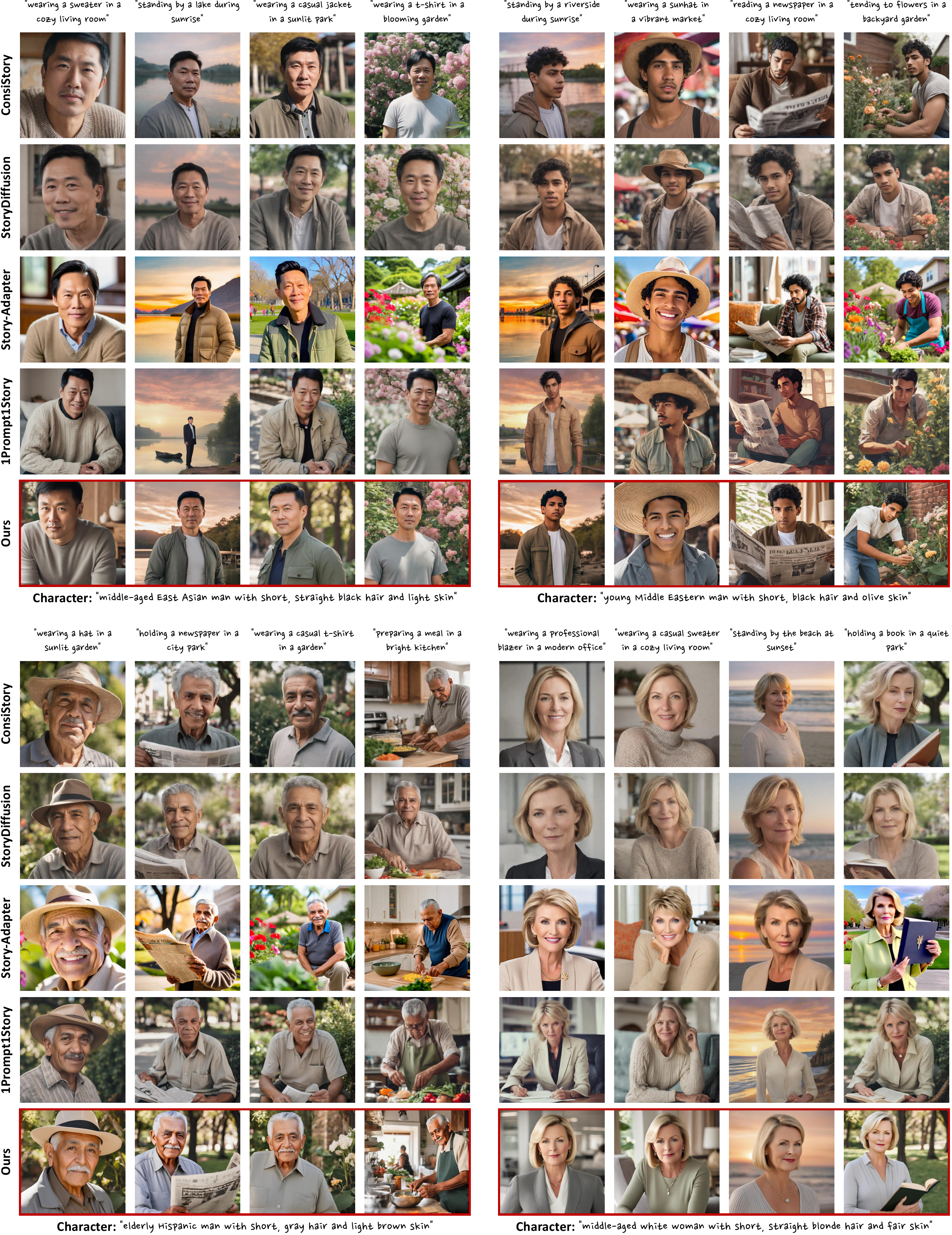}
    
    \vspace{-2mm}
    \caption
    {
        \textbf{More qualitative results.} 
        Compared to others, our \method exhibits remarkable performance in handling human-centric scenarios, enabling consistent generation of human characters with only text as input.
        \textit{Zoom in for better view.}
    }
    \vspace{-5mm}
    
    \label{fig:qual_results_more_1}
     
\end{figure*}

\begin{figure*}[t]
    \centering

    \includegraphics[width=0.95\linewidth]{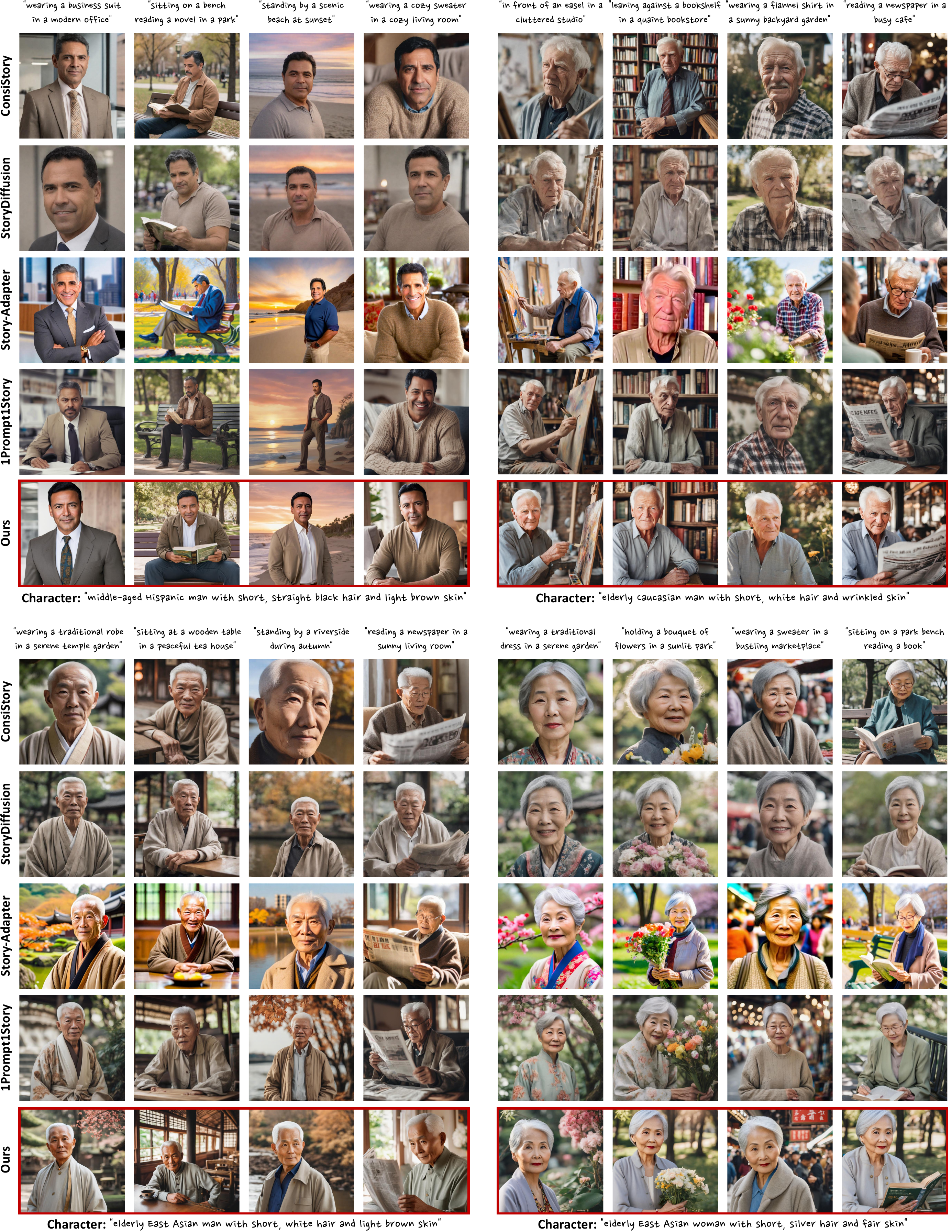}
    
    \vspace{-2mm}
    \caption
    {
        \textbf{More qualitative results.} 
        Compared to others, our \method exhibits remarkable performance in handling human-centric scenarios, enabling consistent generation of human characters with only text as input.
        \textit{Zoom in for better view.}
    }
    \vspace{-5mm}
    
    \label{fig:qual_results_more_2}
     
\end{figure*}

We provide more qualitative results in Figure~\ref{fig:qual_results_more_1} and Figure~\ref{fig:qual_results_more_2}.
Comprehensive comparison shows that our \method can generate high-quality images that depict consistent characters and adhere to text prompts, outperforming existing state-of-the-art methods.


\end{document}